\documentclass[]{interact}

\usepackage{graphicx}%
\usepackage{subfigure}
\usepackage{multirow}%
\usepackage{amsmath,amssymb,amsfonts}%
\usepackage{amsthm}%
\usepackage{mathrsfs}%
\usepackage[title]{appendix}%
\usepackage{xcolor}%
\usepackage{textcomp}%
\usepackage{manyfoot}%
\usepackage{booktabs}%
\usepackage{algorithm}%
\usepackage{algorithmicx}%
\usepackage{algpseudocode}%
\usepackage{listings}%
\usepackage{CJK}
\usepackage{indentfirst}
\usepackage{cases}
\usepackage{longtable}
\usepackage{array}
\usepackage{relsize}
\usepackage{url}

\usepackage{soul} 
\usepackage{color, xcolor} 

\usepackage{epstopdf}
\usepackage[caption=false]{subfig}

\usepackage[numbers,sort&compress]{natbib}
\bibpunct[, ]{[}{]}{,}{n}{,}{,}
\makeatletter
\def\NAT@def@citea{\def\@citea{\NAT@separator}}
\makeatother

\theoremstyle{plain}
\newtheorem{Theorem}{Theorem}
\newtheorem{Corollary}{Corollary}

\theoremstyle{definition}
\newtheorem{Definition}{Definition}
\newtheorem{Example}{Example}

\theoremstyle{remark}
\newtheorem{Remark}{Remark}

\begin{document}
	
	
\title{Bilevel Models for Adversarial Learning and A Case Study}
	
\author{
\name{Yutong Zheng\textsuperscript{a}
and
Qingna Li\textsuperscript{a,b}\thanks{Email: qnl@bit.edu.cn. Corresponding author. } 
}
\affil{\textsuperscript{a}School of Mathematics and Statistics, Beijing Institute of Technology, Beijing, China \textsuperscript{b}Beijing Key Laboratory on MCAACI/Key Laboratory of Mathematical Theory and Computation in Information Security, Beijing Institute of Technology, Beijing, China
}
	}
	
\maketitle
	
\begin{abstract}
Adversarial learning has been attracting more and more attention thanks to the fast development of machine learning and artificial intelligence. However, due to the complicated structure of most machine learning models, the mechanism of adversarial attacks is not well interpreted. How to measure the effect of attacks is still not quite clear. 
		In this paper, we investigate the adversarial learning from the perturbation analysis point of view. 
		We characterize the robustness of learning models through the calmness of the solution mapping. 
		In the case of convex clustering models, we identify the conditions under which the clustering results remain the same under perturbations. 
		When the noise level is large, it leads to an attack. 
		Therefore, we propose two bilevel models for adversarial learning  where the effect of adversarial learning is measured 
		by some deviation function. 
		Specifically, we systematically study the so-called $\delta$-measure and show that under certain conditions, it can be used as a deviation function in adversarial learning for convex clustering models. 
		Finally, we conduct numerical tests to verify the above theoretical results as well as the efficiency of the two proposed bilevel models.
\end{abstract}
	

\section{Introduction}\label{sec1}
With the fast development of machine learning and artificial intelligence, adversarial learning is receiving growing attention. 
The first striking demonstration came from Szegedy et al. \cite{szegedy2014}, who showed that imperceptible perturbations can reliably force modern neural networks to misclassify inputs, thereby exposing a fundamental vulnerability of high-capacity models.
Goodfellow et al. \cite{goodfellow2015} provided a concise explanation and practical attack algorithm so-called the Fast Gradient Sign Method (FGSM) for fast, effective adversarial example synthesis, and thus enabled a set of adversarial-training defenses. 
Kurakin et al. \cite{kurakin2017} showed that adversarial training can be implemented on larger-scale datasets such as ImageNet, and further revealed that this approach leads to a significant improvement in the robustness of one-step methods.
Roberts and Smyth \cite{roberts2022byzantine} analyzed stochastic gradient descent under Byzantine adversaries, highlighting robustness issues that also arise in distributed or large-scale settings.
Chhabra et al. \cite{chhabra2020suspicion} presented a black-box adversarial attack algorithm for clustering models with linearly separable clusters. 
Later, Chhabra et al. \cite{chhabra2022robustness} proposed a black-box adversarial attack against deep clustering models. 
We refer to \cite{yuan2019adversarial,costa2024deep,zhao2025data} for the review and monographs of adversarial learning. 

Following widely adopted taxonomies, adversarial attacks can be divided into three types based on attacker's knowledge: white-box (full access to parameters/gradients), black-box (only queries or input–output pairs), and gray-box in between. Due to different goals of attacks, it can also be classified as confidence reduction, untargeted misclassification and targeted misclassification. 
A large body of work has been devoted to designing powerful white-box attacks such as FGSM \cite{goodfellow2015} and its iterative variants and decision-boundary-based attacks like DeepFool \cite{moosavi2016}.
Feinman et al.\ \cite{feinman2017DetectingAS} proposed to detect adversarial samples using kernel density estimates in the feature space of the last hidden layer together with Bayesian uncertainty estimates.
Ahmed et al.\ \cite{ahmed2024reevaluating} systematically evaluate four representative deep learning attacks against intrusion-detection models on cybersecurity datasets and compare several defense strategies, highlighting how different combinations of attacks and defenses affect the robustness and accuracy of such systems.
Madry et al. \cite{madry2018pgd} cast adversarial training as min–max robust optimization and used multi-step PGD as a strong white-box baseline for $l_p$-bounded threats. Dong et al. \cite{dong2018mifgsm} added momentum to iterative gradients to improve black-box transferability over standard iterative attacks. Brendel et al. \cite{brendel2018decision} proposed boundary attack that needs only top-1 decisions, walking along the boundary while shrinking perturbations.  Papernot et al. \cite{papernot2016jsma} targeted a specific label by perturbing a small, saliency chosen set of pixels (sparse $L_0$-style attack). Eykholt et al. \cite{eykholt2018robust} showed sticker-based perturbations that reliably fool traffic-sign recognition in the real world.

While most exciting literature focuses on developing efficient algorithms to solve adversarial learning model, understanding the mechanism is also extremely crucial in order to improve the robustness of learning models. Below we briefly review some related literature that motivates our work in this paper. 
Moosavi-Dezfooli et al. \cite{moosavi2016} introduced the DeepFool attack, viewing adversarial perturbations as minimal crossings of local decision boundaries. By iteratively linearizing the classifier, it linked perturbations to decision region geometry and provided a principled way to approximate the smallest such perturbations.
Carlini and Wagner \cite{carlini2017} formalized perturbation attacks as optimization problems with tailored loss functions, showing attack strength depends critically on objective function and constraint choices.
Ilyas et al. \cite{ilyas2019} offered a feature-based view, arguing that adversarial perturbations exploit non-robust features: predictive yet human-imperceptible statistical patterns. This shifted the focus from geometry to data representation.
Su, Li and Cui \cite{su2022optimization} systematically studied three types of adversarial perturbations, deriving the explicit solutions for sample-adversarial perturbations (sAP), class-universal adversarial perturbations (cuAP) and universal adversarial perturbations (uAP) for binary classification, and approximate the solution for uAP  multi-classification case.
Later Su and Li \cite{su2022efficient} addressed the difficulty of generating sAP for nonlinear SVMs via implicit mapping by transforming the perturbation optimization into a solvable nonlinear KKT system.

	On the other hand, clustering is a popular yet basic model in machine learning. 
	The study of clustering technology dates back at least to the pioneering work of Driver and Kroeber~\cite{driver1932quantitative}, who quantified cultural relationships by similarity analysis in anthropology, as well as Tryon's monograph on cluster analysis in psychology~\cite{tryon1939cluster}. 
	Since then, many clustering algorithms have been developed, including early agglomerative hierarchical methods based on nearest-neighbour graphs and minimum-variance criteria~\cite{florek1951liaison,ward1963hierarchical}, the K-means algorithm and its modern variants~\cite{macqueen1967some,lloyd1982least},  density-based methods such as DBSCAN \cite{ester1996dbscan} and OPTICS~\cite{ankerst1999optics}, and the more recent density peaks clustering algorithm  (DPC)~\cite{rodriguez2014dpc,wei2023dpc}. We refer the reader to the surveys and monographs~\cite{xu2005survey,everitt2011cluster} for the vast literature on clustering methods and their applications. 
	Below we mainly focus on convex clustering, which is most related to our paper.

Convex clustering model was initially proposed by Pelckmans et al. \cite{pelckmans2005convex}, and was studied in \cite{lindsten2011clustering,hocking2011clusterpath}. The idea of convex clustering model is as follows. If two observations belong to the same cluster, then their corresponding centroids should be the same. 
Convex clustering model \cite{hocking2011clusterpath,lindsten2011clustering,chi2015splitting,sun2025resistant,panahi2017clustering,sun2021convex} has several advantages, such as the uniqueness of solution and theoretical guarantee of cluster recovery. To deal with high dimensional data clustering, Yuan et al. \cite{yuan2022dimension} proposed a dimension reduction technique for structured sparse optimization problems. 
Ma et al. \cite{ma2023improved} proposed an improved robust sparse convex clustering (RSCC) model, which incorporates a novel norm-based feature normalization technique to effectively identify and eliminate outlier features.
Angelidakis et al. \cite{angelidakis2017algorithms} developed improved algorithms for stable instances of clustering problems with center-based objectives, including K-means, K-median and K-center.
However, the adversarial attack case for clustering remains untouched.

To summarize, due to the complicated structure of different learning models, most adversarial attacks are difficult to interpret. Therefore, a natural question is whether we can understand the attack for a simple learning model. This motivates our work. In this paper, we study the mechanism of adversarial attack on clustering models. 
	The contribution of the paper can be summarized as follows.  
	(i) Firstly, we start with the perturbation of learning models, and we qualify the robustness of the solution set of  learning models under perturbation by calmness. By doing so, the robustness of  learning models can be clearly analyzed by the calmness property of the solution mapping. 
	(ii) As a case study, we identify the conditions for the robustness of convex clustering model. That is, under those conditions, the convex clustering model provides unchanged clustering results under perturbation. 
	(iii) We introduce two bilevel optimization models for adversarial learning. 
	In particular, we formulate two bilevel adversarial learning models based on the convex clustering model. 
	(iv) We propose and analyze deviation measures that quantify the impact of adversarial perturbations. 
	In particular, we study the so-called $\delta$-measure and examine its behaviour in a series of 2-way and 3-way clustering examples. 
	We also generalize it to arbitrary K-way clustering. 
	(v) Finally, we verify the above theoretical results and the efficiency of two proposed models by numerical results on the state-of-the-art datasets. 
	The experiments show that convex
	clustering is robust to moderate perturbations, while larger perturbations produce a clear
	staircase behaviour of deviation $\delta$ and $N_{\mathrm{chg}}$. 
    We also compare the direct method and \texttt{fmincon} on the different UCI datasets and find that 
	RI-based function can be used as a reasonable deviation function, whereas NMI can't.

The organization of the paper is as follows. In Section 2, we investigate the perturbation of learning models and relate the sensitivity of solution mapping to the so-called calmness property in the context of perturbation analysis. In Section 3, we study the effect of perturbation on the convex clustering model and provide some  examples. In Section 4, we propose the bilevel optimization models for adversarial learning. In Section 5, we study the so-called $\delta$-measure function to show the well-definedness as the measure of adversarial attack. 
In Section 6, we conduct numerical experiments to verify the theoretical results and the efficiency of bilevel models.
Final conclusions are given in Section 7.

Notations. We use $\|\cdot\|$ as $l_2$ norm for vectors and Frobenius norm for matrices. We use $|V|$ to denote the number of elements in a set $V$.

\section{Perturbation of Learning Models}\label{sec2}
In this part, we will start with the learning model, based on which we will address the perturbation of learning models.

\subsection{Learning Model}
Let $X \subseteq \mathcal{X}$ be the training data, $Y\subseteq \mathcal{Y}$ be the model parameter in a learning model. The learning process (also referred to as training process) is to find the model parameter $	Y^*$ by solving the following model:
\begin{equation}\label{eq-p}
	\min_{Y \in F(X)} L(X, Y)
	\tag{P}
\end{equation}
where $L(X, Y)$ is the objective of the training model, $F(X)\subseteq \mathcal{Y}$ is the feasible set of $Y$, which may be affected by the training data $X$.  Let $Y^*$ be the optimal solution of \eqref{eq-p}, which may not be unique. The solution set of \eqref{eq-p} is denoted as $\mathcal{S}$.

Let the decision function be $D_{Y^*}(\cdot)$, where $D_{Y^*}(x)$ gives the final decision of a new data $x$. For example, for binary classification model, $D_{Y^*}(x)$ is the sign function which gives the label of data $x$, by applying the learning result $Y^*$. Specifically, we give two simple examples below.

\begin{Example}\label{eg-svm}
	Support Vector Machine (SVM) for binary classification  \cite{cortes1995support}, where
	$X = \left[ \begin{pmatrix} x_1 \\ y_1 \end{pmatrix}, \cdots, \begin{pmatrix} x_n \\ y_n \end{pmatrix} \right] \in \mathbb{R}^{(d+1)\times n},\ Y=(\omega, b)\in  \mathbb{R}^{d+1}$, with the learning model
	\begin{equation}\label{eq-l2svm}
		\min_{(\omega, b)\in  \mathbb{R}^{d+1}} \quad \frac{1}{2} \|\omega\|^2 + C \sum_{i=1}^{n} \left( \max\left(0, 1 - y_i(\omega^\top {x}_i + b)\right) \right)^2:=L^{SVM}(X,Y).
		\tag{$l_2$-SVM}
	\end{equation}
	The optimal solution of \eqref{eq-l2svm} is denoted as $(\omega^*,b^*)$. The decision function $D^{SVM}_{(\omega^*,b^*)}(x)=\text{sign}({\omega^*}^\top x+b^*)\in \{-1, 1\}$.
	
\end{Example}

\begin{Example}\label{eg-cc}
	Convex Clustering \cite{sun2021convex}, where $X = [x_1, \cdots, x_n] \in \mathbb{R}^{d \times n}$, $Y = [y_1, \cdots, y_n] \in \mathbb{R}^{d \times n}$ with the learning model
	\begin{equation}\label{eq-cvc}
		\min\limits_{Y \in \mathbb{R}^{d \times n}}L^{CVC}(X,Y) = \frac{1}{2}\sum\limits_{i=1}^n \|y_i - x_i\|^2 + \gamma \sum\limits_{1 \leq i < j \leq n} w_{ij} \|y_i - y_j\|_p
		\tag{CVC}
	\end{equation}
	where $\gamma>0,\ w_{ij}\geq0,\ i,j=1,\cdots,n$ are given.
	The optimal solution of \eqref{eq-cvc} is $Y^* = [y_1^*, \cdots, y_n^*] \in \mathbb{R}^{d \times n}$.
	The decision function (i.e., the clustering result) is given by
	\begin{equation}\label{eq-dmap}
		D^{CVC}_{Y^*}= \{ V_1, \cdots, V_K\} ,
	\end{equation}
	where $ \{ V_1, \cdots, V_K\}$ is a partition of $\{1,\cdots,n\}$ and $x_i$ and $x_j$ are in the same partition if $y_i^* = y_j^*$. 	Since $L^{CVC}(\cdot,\cdot)$ is also strongly convex in $Y$, $Y^*$ is the unique solution of \eqref{eq-cvc}. That is, $S$ is a single singleton.
\end{Example}

\begin{Remark}
	For most complicated learning models such as convolutional neural network (CNN), it is usually difficult to write down an explicit objective function $L(X,Y)$ and the associated decision function $D_{Y^*}$. This makes it challenging both to solve the corresponding bilevel adversarial learning models and to interpret the role of adversarial perturbations in a precise way. 
\end{Remark}

\subsection{Perturbation of Learning Models}\label{subsec-learningmodel}
Having introduced the learning model, we are ready to consider the perturbation of the learning model. For general optimization problems, the perturbation analysis is fully addressed in \cite{bonnans2013perturbation}. 
In the case of noise, let $X(\varepsilon)$ be the noised training data set where $\varepsilon \in \mathcal{X}$ is the perturbation (or noise), the output of the learning model under the perturbation $\varepsilon$ is as follows. 
\begin{equation}\label{eq-pvar}
	\min_{Y \in F(X(\varepsilon))} L(X(\varepsilon), Y)
	\tag{P$_\varepsilon$}
\end{equation}
where $ F(X(\varepsilon))\subseteq \mathcal{Y}$ is the feasible set of $Y$, which may be affected by the noised training data $X(\varepsilon)$.
Let $Y^*(\varepsilon)$ denotes the optimal solution of the  learning model under perturbation $\varepsilon$ and $\mathcal{S}(\varepsilon)$ denotes the solution set of \eqref{eq-pvar}.

It is obvious that $X(0) = X$ and $S(0)=S$.  
Here we would like to highlight that due to different learning models, the form of $\varepsilon$ could be different. If $\mathcal{X} = \mathbb{R}^{k \times n}$, then $\varepsilon$ could be the additive noise or multiplicative noise. 
The noise can also be point-wise modifications or feature-level modifications. 
If $\mathcal{X}$ is in the graph space in graph clustering, i.e., 
$
\mathcal{X} = \left\{ \mathcal{G}= (V_{\mathcal{G}}, E)\ |\  |V_{\mathcal{G}}| = n,\ E \text{ is any set of edges on } V_{\mathcal{G}} \right\}
$
, then $X(\varepsilon)$ could be the graph-based perturbation that removing vertices or changing some edges from the current graph $X$.

	Intuitively, if the perturbation $\varepsilon$ on data is relatively small, $Y^*(\varepsilon)$ may still be different from $Y^*$ \footnote{Note that $Y^*(0)=Y^*$. For simplicity, we always use $Y^*$ instead of $Y^*(0)$.},
	leading to possibly small changes in $S(\varepsilon)$ compared to $S$. If we take $S(\varepsilon)$ as a multifunction of $\varepsilon$, 
	one way to measure the changes of the solution set $S(\varepsilon)$ under the perturbation $\varepsilon$ is calmness, which is a useful property in perturbation analysis. We give the definition below.

	\begin{Definition}[Calmness]\cite[Definition 2]{flegel2007optimality}\label{def:calmness}
		Let $\mathcal{S}(\varepsilon)=\arg\min_{Y\in F(X(\varepsilon))} L(X(\varepsilon),Y)$ be a multifunction with a closed graph, denoted as $\operatorname{gph} \mathcal{S}(\varepsilon)$, and $(\bar\varepsilon,\overline Y)\in\operatorname{gph} \mathcal{S}(\varepsilon)$. We say that $\mathcal{S}(\cdot)$ is calm at $(\bar\varepsilon,\overline Y)$ provided that there exist neighborhoods $\mathcal N_{\varepsilon}$ of $\bar\varepsilon$ and $\mathcal N_{Y}$ of $\overline Y$, and a modulus $\mathcal{L}\ge 0$, such that
	\[
	\mathcal{S}(\varepsilon)\cap \mathcal N_{Y}\subset \mathcal{S}(\bar\varepsilon)+\mathcal{L}\|\varepsilon-\bar\varepsilon\|\\ \mathbb{B}
	\quad\text{for all } \varepsilon\in\mathcal N_{\varepsilon},
	\]
	where $\mathbb B$ denotes the closed unit ball in $\mathcal{X}$ and $\operatorname{gph} \mathcal{S}(\varepsilon):=\{(\varepsilon,Y): Y\in S(\varepsilon)\}$.
	\end{Definition}

	Based on the definition of calmness, one can see that the calmness of $\mathcal{S}(\cdot)$ at $\varepsilon=0$ is particularly useful for measuring the changes of $\mathcal{S}(\varepsilon)$ relative to $\mathcal{S}(0)$. We formally give it below.
	
	\begin{Definition}[Calmness at ${\varepsilon=0}$]
		$\mathcal{S}(\cdot)$ is calm at $(0,\overline Y)$ if there exist neighborhoods $\mathcal N_{\varepsilon}$ of $0$, $\mathcal N_{Y}$ of $\overline  Y$, and a modulus $\mathcal{L}_0>0$ such that
		\[
		\mathcal{S}(\varepsilon)\cap \mathcal N_{Y}\subset \mathcal{S} +\mathcal{L}_0\|\varepsilon\|\ \mathbb B
		\quad\text{for all } \varepsilon\in\mathcal N_{\varepsilon}.
		\]
	\end{Definition}

	Therefore, one can see that if $\mathcal{S}(\cdot)$ is calm at $0$, then the changes in the solution set $\mathcal{S}(\varepsilon)$ can be controlled by the changes in $\|\varepsilon\|$ (up to the scalar $\mathcal{L}_0$). In other words, the robustness of the learning model $L(\cdot,\cdot)$ is closely related to the calmness of the solution set $\mathcal{S}(\cdot)$. For a set $\mathcal{S}(\cdot)$, there are various ways to check calmness \cite{gfrerer2011subregcalm,kruger2015holderSubreg,gfrerer2016implicit}; moreover, Zhou and So \cite{zhou2017unified} provided an equivalent characterization by showing that the error bound property holds if and only if a suitably defined solution mapping  is calm, and we will not discuss the details here.
	
    Having successfully measured the changes of $\mathcal{S}(\varepsilon)$ due to the perturbation $\varepsilon$, we move on to see whether there is any change in the decision function $D_{Y^*}(\cdot)$. Even when the solution set $\mathcal{S}(\varepsilon)$ changes, it is still possible that the decision function remains the same.
    The reason is as follows. In many learning tasks, the decision function is discontinuous. For example, binary classification as shown in Example \ref{eg-svm}. 
In other words, for such situation, the perturbation $\varepsilon$ does not have effect on the learning result. We can regard $\varepsilon$ as a neglectable noise in this case. Therefore, the question we would like to ask is as follows: under what condition on $\varepsilon$, the decision function is not changed, i.e., $D_{Y^*(\varepsilon)}= D_{Y^*}$? 
	The question is not easy to answer. As we mentioned before, this is also highly related to the specific form of the learning model. We will address this question in Section 3, by looking at the convex clustering model as an example.
	
\section{Perturbation Analysis for Convex Clustering}\label{sec3}
In this section, we take clustering as an example to study the effect of perturbation. 
	We choose the convex clustering model due to the following reasons. Firstly, the strong convexity of $L^{CVC}(\cdot)$ guarantees the unique solution of \eqref{eq-cvc}, that is, $\mathcal{S}(\varepsilon)$ is a singleton for each $\varepsilon$. 
	Secondly, there are exact recovery theoretical results under proper assumptions, which states that under some conditions, the solution of \eqref{eq-cvc} perfectly recovers the ground truth clustering. 
	Finally, Ssnal \cite{sun2021convex} was proposed to solve \eqref{eq-cvc} which is proved to be highly efficient. We will use it to solve \eqref{eq-cvc} in our numerical experiments. 
We start with the case where the small perturbation will not change the clustering result.

Let $X=[x_1,\cdots,x_n]$ be the data and $\mathcal{V}= \{ V_1, \dots, V_K \}$ be a partitioning of $X$ and $K$ is the number of clusters. 
The index sets are defined by 

\begin{equation*}
	\begin{aligned}
		&I_\alpha:=\left\{i \mid {x}_i \in V_\alpha\right\},\ n_\alpha=|I_\alpha|,\  \text { for } \alpha=1,2, \ldots, K ,\\
		& {x}^{(\alpha)}=\frac{1}{n_\alpha} \sum_{i \in I_\alpha} {x}_i, \quad w^{(\alpha, \beta)}=\sum_{i \in I_\alpha} \sum_{j \in I_\beta} w_{i j}, \quad \forall \alpha, \beta=1, \ldots, K, \\
		& w_i^{(\beta)}=\sum_{j \in I_\beta} w_{i j}, \quad \forall i=1, \ldots, n,\  \beta=1, \ldots, K .
	\end{aligned}
\end{equation*}
The following result shows the exact recovery result of the learning model \eqref{eq-cvc}.
\begin{Theorem} \label{thm5insun}
	\cite[Theorem 5]{sun2021convex}
	Consider the input data $X=\left[{x}_1, \cdots, {x}_n\right]\in \mathbb{R}^{d \times n}$ and its partitioning $\mathcal{V}=\left\{V_1, V_2, \ldots, V_K\right\}$. Assume that all centroids $\left\{{x}^{(1)}, {x}^{(2)}, \ldots, {x}^{(K)}\right\}$ are distinct. Let $q \geq 1$ be the conjugate index of $p$ such that $\frac{1}{p}+\frac{1}{q}=1$. $Y^*$ is the unique solution of \eqref{eq-cvc} and define the map $f\left({x}_i\right)={y}_i^*$ for $i=1, \cdots, n$.
	Let
	\[ \mu_{i j}^{(\alpha)}:=\sum_{\beta=1,\  \beta \neq \alpha}^K\left|w_i^{(\beta)}-w_j^{(\beta)}\right|, \quad i, j \in I_\alpha,\  \alpha=1,2, \ldots, K . \]
	Assume that 
	\begin{equation}\label{eq-sun-con1}
		w_{i j}>0\ \text{and}\ n_\alpha w_{i j}>\mu_{i j}^{(\alpha)}\ \text{for\ all}\ i, j \in I_\alpha,\ \alpha=1, \ldots, K. 
		\tag{C1}
	\end{equation}
	Let
		\small
	\begin{equation*}
		\gamma_{\min }  :=\max _{1 \leq \alpha \leq K} \max _{i, j \in I_\alpha}\left\{\frac{\left\|{x}_i-{x}_j\right\|_q}{n_\alpha w_{i j}-\mu_{i j}^{(\alpha)}}\right\}, \ \ 	
		\gamma_{\max } :=\min _{1 \leq \alpha<\beta \leq K}\left\{\frac{\left\|{x}^{(\alpha)}-{x}^{(\beta)}\right\|_q}{\frac{1}{n_\alpha} \sum\limits_{1 \leq l \leq K, l \neq \alpha} w^{(\alpha, l)}+\frac{1}{n_\beta} \sum\limits_{1 \leq l \leq K, l \neq \beta} w^{(\beta, l)}}\right\} .
	\end{equation*}
	\normalsize
	If 
	\begin{equation}\label{eq-sun-con2}
		\gamma_{\text{min}} < \gamma_{\text{max}}
		\tag{C2}
	\end{equation}
	and $\gamma$ is chosen such that $\gamma \in\left[\gamma_{\min }, \gamma_{\max }\right)$, then the map $f$ perfectly recovers $\mathcal{V}$.
\end{Theorem}
Theorem \ref{thm5insun} shows that if conditions \eqref{eq-sun-con1} and \eqref{eq-sun-con2} hold, then the clustering result $D^{CVC}_{Y^*}$ coincides with the ground truth partition of $X$. That is, $D^{CVC}_{Y^*}=\mathcal{V}$.
One can see that the exact recovery is based on conditions \eqref{eq-sun-con1} and \eqref{eq-sun-con2}. Moreover, \eqref{eq-sun-con1} and \eqref{eq-sun-con2} are calculated based on the ground truth partition $\mathcal{V}$ as well as the data $X$. Given the perturbed data $X(\varepsilon)$, one can make use of Theorem \ref{thm5insun} and provide a sufficient condition under which the clustering result is unchanged under perturbation $\varepsilon$.
To that end, let $X(\varepsilon)=[x_1(\varepsilon),\cdots,x_n(\varepsilon)]$ be the perturbation of $X \in \mathbb{R}^{d \times n}$ with $\varepsilon \in \mathbb{R}^{d \times n}$. Under the partition of $D^{CVC}_{Y^*}=\left\{V_1, V_2, \ldots, V_K\right\}$, we define the following notations
\begin{equation}\label{eq-index}
	\begin{aligned}
		& {x}^{(\alpha)}({\varepsilon})=\frac{1}{n_\alpha} \sum_{i \in I_\alpha} {x}_i(\varepsilon), \quad w^{(\alpha,\beta)}({\varepsilon})=\sum_{i \in I_\alpha} \sum_{j \in I_\beta} w_{i j}({\varepsilon}), \quad \forall \alpha, \beta=1, \ldots, K ,\\
		& w_i^{(\beta)}({\varepsilon})=\sum_{j \in I_\beta} w_{i j}({\varepsilon}), \quad \forall i=1, \ldots, n,\  \beta=1, \ldots, K .
	\end{aligned}
\end{equation}
Here we use ${x}^{(\alpha)}({\varepsilon}),\ w^{(\alpha,\beta)}({\varepsilon})$ and $w_i^{(\beta)}({\varepsilon})$ to mean that those coefficients may be related to the perturbation $\varepsilon$.

\begin{Theorem}\label{thm1}
	Consider the perturbed data $X(\varepsilon) = [x_1(\varepsilon), \dots, x_n(\varepsilon)] \in \mathbb{R}^{d \times n}$ and the partitioning  $D^{CVC}_{Y^*}=\left\{V_1, V_2, \ldots, V_K\right\}$. 
	Let ${x}^{(\alpha)}({\varepsilon})$, $ w^{(\alpha, \beta)}({\varepsilon})$ and $w_i^{(\beta)}({\varepsilon})$ be defined as in \eqref{eq-index}.
	Assume that all centroids $\left\{ x^{(1)}(\varepsilon), \dots, x^{(K)}(\varepsilon)\right\}$ are distinct. Let $q \geq 1$ be the conjugate index of $p$ such that $\frac{1}{p} + \frac{1}{q} = 1$.
	Let $Y^*(\varepsilon) = [y^*_1(\varepsilon), \dots, y^*_n(\varepsilon)]$ be learned via \eqref{eq-cvc} in  Example \ref{eg-cc} and $f_{\varepsilon}: X(\varepsilon) \to Y^*(\varepsilon)$ is given by $f_{\varepsilon}(x_i\left(\varepsilon)\right) = y^*_i(\varepsilon)$.
	Let $\mu_{ij}^{(\alpha)}({\varepsilon}) := \sum\limits_{\beta = 1, \beta \neq \alpha}^k \left|w_i^{(\beta)}({\varepsilon}) - w_j^{(\beta)}({\varepsilon})\right|,\ i, j \in I_\alpha,\ \alpha = 1, \dots, K$.
	Assume that 
	\begin{equation}\label{eq-cond1}
		w_{ij}(\varepsilon) > 0\  \text{and}\ n_\alpha w_{ij}(\varepsilon) > \mu_{ij}^{(\alpha)}(\varepsilon)\  \text{for all}\  i, j \in I_\alpha,\ \alpha = 1, \dots, K.   
		\tag{C1$^{\prime}$}
	\end{equation}
	Let 
	\begin{equation*}
		\gamma_{\text{min}}^{\varepsilon} := \max\limits_{1 \leq \alpha \leq K} \max\limits_{i,j \in I_{\alpha}} \left\{ \frac{\| x_i(\varepsilon) - x_j(\varepsilon) \|_q}{n_\alpha w_{ij}(\varepsilon) - \mu_{ij}^{(\alpha)}(\varepsilon)} \right\},
	\end{equation*}
	\begin{equation*}
		\gamma_{\text{max}}^{\varepsilon} := \min\limits_{1 \leq \alpha<\beta \leq K} \left\{ \frac{\| {x}^{(\alpha)}(\varepsilon) - {x}^{(\beta)}(\varepsilon) \|_q} {\frac{1}{n_\alpha} \sum\limits_{1 \leq l \leq K, l \neq \alpha} w^{(\alpha, l)}(\varepsilon)+\frac{1}{n_\beta} \sum\limits_{1 \leq l \leq K, l \neq \beta} w^{(\beta, l)}(\varepsilon)}\right\}.
	\end{equation*}
	If 
	\begin{equation}\label{eq-cond2}
		\gamma_{\text{min}}^{\varepsilon} < \gamma_{\text{max}}^{\varepsilon}
		\tag{C2$^{\prime}$}
	\end{equation}
	and $\gamma$ is chosen such that $\gamma\in [\gamma_{\text{min}}^{\varepsilon}, \gamma_{\text{max}}^{\varepsilon})$, then $D^{CVC}_{Y^*(\varepsilon)}=D^{CVC}_{Y^*}$, i.e., the clustering result is unchanged.
\end{Theorem}

\textbf{Proof.} 
	By applying \cite[Theorem 5 ]{sun2021convex} with the ground truth partitioning $\mathcal{V} = D^{CVC}_{Y^*}$, and $X$ replaced by $X(\varepsilon)$, we get that the mapping $f_{\varepsilon}: X(\varepsilon) \rightarrow Y^*(\varepsilon)$ recovers the partitioning $\mathcal{V}$. That is, $D^{CVC}_{Y^*(\varepsilon)}$ is the same as $D^{CVC}_{Y^*}$. The proof is finished.
   \hfill$\square$

We demonstrate this by the following one-dimension example with two clusters, that is, $d=1$ and $K=2$. We use the weighted matrix $W=(w_{ij})=E_n$\ ( $E_{n_1}$ denotes the matrix of size $n_1\times n_1$ whose elements are all ones) and $p=2$. We only consider adding perturbation to a specific data. The solution of \eqref{eq-cvc} is obtained by running the algorithm semismooth Newton-CG augmented Lagrangian method (Ssnal) \footnote{https://www.polyu.edu.hk/ama/profile/dfsun//Codes/Statistical-Optimization/} in \cite{sun2021convex}.

\begin{figure}[H]
	\centering
	\includegraphics[width=0.8\textwidth, keepaspectratio]{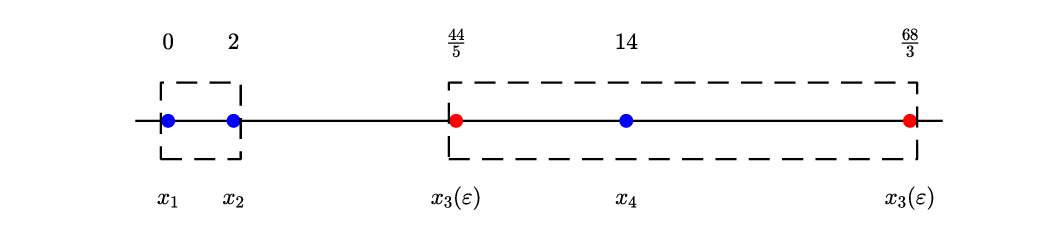}
	\caption{
		Perturbed dataset $X(\varepsilon)$ with $x_3(\varepsilon) \in \left(\tfrac{44}{5}, \tfrac{68}{3}\right)$, where the conditions \eqref{eq-cond1} and \eqref{eq-cond2} hold and the clustering results remain unchanged.}
	\label{fig:eg3-2}
\end{figure}

\begin{Example}\label{eg1}
	Let $X=[0,\ 2,\ 10,\ 14]\in\mathbb{R}^{1 \times 4}$, as shown in Figure \ref{fig:eg3-1}. The solution of convex clustering model in \eqref{eq-cvc} is $Y^{*}=[1,\ 1,\ 12,\ 12]$, with decision function $\mathcal{V}=D^{CVC}_{Y^*}=\left\{ \{1,2\},\{3,4\}\right\}. $
	\begin{figure}[H]
		\centering
		\includegraphics[width=0.8\textwidth, keepaspectratio]{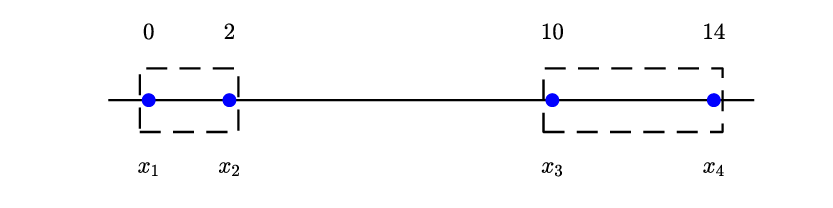}
		\caption{The original one-dimensional dataset $X=[0,\ 2,\ 10,\ 14]$ to illustrate convex clustering and its robustness to data perturbations.}
		\label{fig:eg3-1}
	\end{figure}
	We perturb only on $x_3$. Let $X(\varepsilon) = [0,\ 2,\ 17,\ 14]$, that is, $\varepsilon = [0,\ 0,\ 7,\ 0]$. Easily see that $\mu_{12}^{(1)}(\varepsilon)=0,\  \mu_{34}^{(2)}(\varepsilon)=0$. 	
	Moreover, $n_\alpha w_{ij}(\varepsilon)-\mu_{ij}^{(\alpha)}(\varepsilon)=n_\alpha>0,\  \text{for all}\  i, j \in I_\alpha,\ \alpha = 1,2$, and $\gamma_{\min }^{\varepsilon}=\max\left\{  \frac{2}{2},\frac{3}{2}\right\} = \frac{3}{2}<\gamma_{\max }^{\varepsilon}=\frac{14.5}{4}=\frac{29}{8}$.
	Therefore, conditions \eqref{eq-cond1} and \eqref{eq-cond2} hold and $D^{CVC}_{Y^*(\varepsilon)} = D^{CVC}_{Y^*} $. In fact, the solution of \eqref{eq-cvc} gives $Y^*(\varepsilon)=[1,\ 1,\ 15.5,\  15.5]$.
\end{Example}
Following Example \ref{eg1}, we can similarly calculate that for any $\varepsilon_3 \in \left(-\frac{6}{5}, \frac{38}{3}\right)$, that is, $x_3(\varepsilon) \in \left(\frac{44}{5}, \frac{68}{3}\right)$, conditions \eqref{eq-cond1} and \eqref{eq-cond2} both hold, implying that the clustering results will not be changed. This is indeed the truth since one can verify it by eyesight (See Figure \ref{fig:eg3-2}).

In fact, if $\varepsilon$ does not satisfy \eqref{eq-cond1} or \eqref{eq-cond2}, it is very likely that the clustering result will change compared with the unperturbed clustering result. Below we give another example to show this phenomenon.

\begin{Example}
	Let $X=[0,\ 2,\ 10,\ 14],\ \mathcal{V}=D^{CVC}_{Y^*}=\left\{ \{1,2\},\{3,4\}\right\} $. Let $X(\varepsilon)=[0,\ 2,\ -4,\ 14]$ with $\varepsilon = [0, 0, -14, 0]$. One can see that $\mu_{12}^{(1)}(\varepsilon)=0,\  \mu_{34}^{(2)}(\varepsilon)=0$, implying that $n_\alpha w_{ij}(\varepsilon)-u_{ij}^{(\alpha)}(\varepsilon)>0,\  \text{for all}\  i, j \in I_\alpha,\ \alpha = 1,2$. However, $\gamma_{\min }^{\varepsilon}=\max\left\{  \frac{2}{2},\frac{18}{2}\right\} =9>\gamma_{\max }^{\varepsilon}=\frac{4}{4}=1$. That is, condition \eqref{eq-cond2} fails. In fact, $Y^{*}(\varepsilon)=[-0.6667,\ -0.6667,\ -0.6667,\ 14]$, which gives $D^{CVC}_{Y^*(\varepsilon)}=\left\{ \{1,2,3\},\{4\}\right\}$. Indeed, it can be noticed from Figure \ref{fig:eg4} that $D^{CVC}_{Y^*(\varepsilon)}\neq D^{CVC}_{Y^*}$.
	\begin{figure}[H]
		\centering
		\includegraphics[width=0.8\textwidth, keepaspectratio]{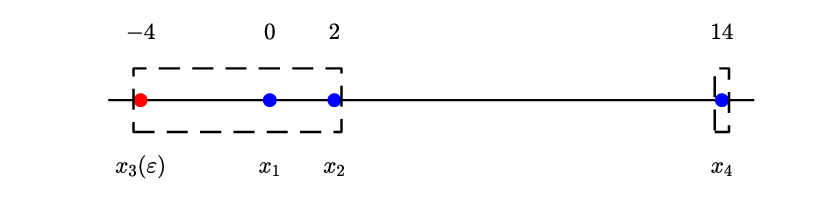}
		\caption{
			Perturbed dataset $X(\varepsilon) = [0,\ 2,\ -4,\ 14]$, where condition \eqref{eq-cond2} fails and the convex clustering result changes from $\left\{ \{1,2\},\{3,4\}\right\}$ to  $\left\{ \{1,2,3\},\{4\}\right\}$.
			}
		\label{fig:eg4}
	\end{figure}
	
\end{Example}

The above example gives rise to another interesting question: how to choose $\varepsilon$ in order to make the clustering result change? In fact, we have the following necessary condition for changing the clustering results. 
\begin{Corollary}\label{thm-nifou}
	If $D^{CVC}_{Y^*(\varepsilon)}\neq D^{CVC}_{Y^*}$, then either \eqref{eq-cond1} or \eqref{eq-cond2} fails for perturbed data $X(\varepsilon)$ under  partition $D^{CVC}_{Y^*}$.
\end{Corollary}
\textbf{Proof.} Assume for contradiction that $\varepsilon$ satisfies both \eqref{eq-cond1} and \eqref{eq-cond2} with perturbed data $X(\varepsilon)$ under  partition $D^{CVC}_{Y^*}$. By Theorem \ref{thm1}, it holds that $D^{CVC}_{Y^*(\varepsilon)}=D^{CVC}_{Y^*}$, implying that the clustering results for $X(\varepsilon)$ will remain the same as $X$, which is a contradiction. Therefore, the proof is finished. \hfill$\square$

\vspace{3mm}
Here we would like to highlight the following two important issues. 
\begin{itemize}
   \item [(i)] 
   	The assumptions of distinct centroids and strictly positive weights, which
   	appear in Theorem~\ref{thm1}, are inherited from the
   	recovery result of convex clustering \cite[Theorem 5]{sun2021convex} and are standard in that context.
   	These conditions ensure
   	that the convex clustering model admits a unique, well-separated partition and
   	allow us to derive clean sufficient conditions under which this partition is preserved under perturbations. For practical clustering problems, these assumptions may fail, putting the exact recovery in question. Therefore, it would be interesting to further study the exact recovery result when the above assumptions fail. We leave this interesting question as our future research topic to study. 
 \item [(ii)] As mentioned in Section \ref{subsec-learningmodel}, the perturbation of $X(\varepsilon)$ can be in various forms. The results in Theorem \ref{thm1} and Corollary \ref{thm-nifou} hold not only for the additive point-wise noise, but also hold for other noise situations, such as multiplicative noise, feature-level noise. 
\end{itemize}

\vspace{3mm}
To conclude this section, we give a short summary. We discussed the role of $\varepsilon$ in the perturbation of convex clustering. In Theorem \ref{thm1}, we identified conditions on $\varepsilon$ under which the clustering result remains the same. We also provided a necessary condition on $\varepsilon$ in order that the clustering result can be changed. 

\section{Bilevel Model for Adversarial Learning}\label{sec4}
In this part, we will reformulate adversarial learning by bilevel optimization models. For general modeling and algorithmic background on bilevel optimization, see  \cite{kleinert2021survey,chen2025set}. 

As we mentioned in Section \ref{sec3}, the perturbation $\varepsilon$ on $X$ may lead to the changed decision function, which means that the learning model is robust. On the other hand, the perturbation $\varepsilon$ on $X$ may lead to the change in decision result $D_{Y^*(\varepsilon)}$. It has two implications. Firstly, it means that such kind of noise is not neglectable. One needs to either do the denoising process to get rid of such noise or to improve the training process to make the learning model more robust. Secondly, from data attacking point of view, attack happens in such situation. In this case, a question arises: how could we choose the perturbation $\varepsilon$ such that some attacking criteria is maximized or minimized? This leads to the following two models for adversarial learning. 
The first model is to make the output of the perturbed learning model as huge difference as possible compared to the original learning model, which is described as ($a>0$ is given)
\begin{equation}\label{eq-adv1}
	\begin{split}
		\max_{\varepsilon \in \mathcal{X}} &\ U\left(\varepsilon\right) \\
		\text{s.t. } &\ Y^*(\varepsilon) \in \mathcal{S}(\varepsilon),\\  
		&\ \|\varepsilon\| \leq a,
	\end{split}
	\tag{BL$_1$}
\end{equation}
where $U(\cdot): \mathcal{X} \to \mathbb{R}$ is defined to be the deviation  function which measures the effect of attack, that is, the changes in decision $D_{Y^*(\varepsilon)}$ compared with the  decision $D_{Y^*}$. To make $U(\cdot)$ well representing the effect of attack, $U(\cdot)$ must have the following properties:
\begin{itemize}
	\item [(i)] $U\left(\varepsilon\right)$ should be a nondecreasing function with respect to $\|\varepsilon\|$. 
	\item [(ii)] $U\left(\varepsilon\right)\geq0$ for any $\varepsilon \in \mathcal{X}$, in particular, $U\left(\varepsilon\right)>0$ if $\varepsilon\neq0$.
	\item [(iii)] $U\left(0\right)=0$. That is,  if there is no perturbation ($\varepsilon=0$), the deviation should be zero.
\end{itemize}
In \eqref{eq-adv1}, the constraint $\|\varepsilon\| \leq a$ controls the magnitude of $\varepsilon$, so that the perturbation is not too large. Otherwise, it would be easily detected and the attack would fail. In practice, the threshold $a$ can be calibrated using prior knowledge on admissible noise levels in a given application. For example, many clustering robustness studies explicitly test algorithms under injected noise ratios around $15\%$ to $30\%$, such as adding $30\%$ uniform or Gaussian noise to the data \cite{li2007noise,zhang2016large}, and one may choose $a$ in accordance with such empirical noise levels. 
One can see that \eqref{eq-adv1} is a bilevel optimization problem where $Y^*(\varepsilon)\in \mathcal{S}(\varepsilon)$ describes the lower level problem, saying that $Y^*(\varepsilon)$ must be the solution of \eqref{eq-pvar}.

The second model is to minimize the scale of perturbation $\varepsilon$, such that the attacking effect reaches the prescribed effect level $\delta_0>0$. That is,
\begin{equation}\label{eq-adv2}
	\begin{split}
		\min_{\varepsilon \in \mathcal{X}} &\ \|\varepsilon\| \\
		\text{s.t. } &\ Y^*(\varepsilon) \in \mathcal{S}(\varepsilon), \\
		&\ U\left(\varepsilon\right) \geq \delta_0.
	\end{split}
	\tag{BL$_2$}
\end{equation}

\begin{Remark}
	$U(\cdot)$ is usually difficult to design due to different learning models. As far as we know, there is little work focusing on developing efficient deviation function, which is in fact important to the two adversarial learning models. We will address this question in the scenario of clustering, which can be found in Section \ref{sec5}.
\end{Remark}

	Bilevel models \eqref{eq-adv1} and \eqref{eq-adv2} are flexible to accommodate many concrete learning tasks. 
	At the lower level, one may take
	$\ell_2$-regularized least squares, logistic regression, support vector machines, or deep neural networks \cite{franceschi2018bilevel,mackay2019hyperparameter}. 
	The deviation function $U(\cdot)$ may take prediction loss on
	a validation set or robust loss in adversarial training
	\cite{madry2018towards,zhang2022fastat}. 
	Moreover, $U(\cdot)$ may also represent robustness-oriented criteria inspired
	by risk-driven anti-clustering, such as the risk-weighted diversity objective
	used to partition training data in~\cite{mauri2023riskanticlustering}. 
	Exploring these deviation choices within our bilevel models thus offers several promising directions for future research.

	Various approaches can be explored to solve the adversarial model \eqref{eq-adv1}. 
	For the white-box setting, where the lower-level model is fully accessible, many approaches can be employed, such as 
   the KKT-based approaches~\cite{dempe2002foundations,colson2007overview},   
    value-function-based 
    approaches~\cite{dempe2013value,gao2022value}, 
    the duality-based approaches~\cite{ouattara2018duality,li2023novel} 
    and the hypergradients-based approaches~\cite{andrychowicz2016learning,maclaurin2015gradient,finn2017model}. 
	On the other hand, the bilevel model also applies to black-box attacks. 
	In such cases, the lower-level model does not admit explicit mathematical formula, one may use evolutionary strategies \cite{wierstra2014natural,hansen2016cma} or derivative-free methods \cite{zhang2024prima,li2011class} to solve \eqref{eq-adv1}.

	Comparing \eqref{eq-adv2} with \eqref{eq-adv1}, \eqref{eq-adv2} is usually more challenging to solve. The reason is that the feasible region of $U\left(\varepsilon\right) \geq \delta_0$ is difficult to explicitly represent due to the implicit form of $Y^*(\varepsilon)$. Therefore, how to design efficient algorithms to solve \eqref{eq-adv2} is an interesting topic which is worth further investigation.

 For convex clustering problem, bilevel models \eqref{eq-adv1} and \eqref{eq-adv2} reduce to the following form 
 \begin{equation}\label{eq-adv1_cvc}
 	\begin{split}
 		\max_{\varepsilon \in \mathcal{X}} &\ U\left(\varepsilon\right) \\
 		\text{s.t. } &\ Y^*(\varepsilon) \in \arg\min\limits_{Y \in \mathbb{R}^{d \times n}}L^{CVC}(X(\varepsilon),Y),\\  
 		&\ \|\varepsilon\| \leq a,
 	\end{split}
 	\tag{BL$_1^{CVC}$}
 \end{equation}
 \begin{equation}\label{eq-adv2_cvc}
 	\begin{split}
 		\min_{\varepsilon \in \mathcal{X}} &\ \|\varepsilon\| \\
 		\text{s.t. } &\ Y^*(\varepsilon) \in \arg\min\limits_{Y \in \mathbb{R}^{d \times n}}L^{CVC}(X(\varepsilon),Y), \\
 		&\ U\left(\varepsilon\right) \geq \delta_0.
 	\end{split}
 	\tag{BL$_2^{CVC}$}
 \end{equation}
We will discuss how to choose $U(\cdot)$ for clustering in the following section.

\section{A Case Study on Deviation Functions}\label{sec5}
In this part, we will study the $\delta$-measure function in adversarial learning for clustering problems, to verify whether it is a deviation function.

As we mentioned above, due to the different structures of learning models as well as the variants of decision functions, the deviation function on the decision function can be in many different forms. 
Taking the binary classification as an example, the deviation function can be chosen as the norm of difference of the classification results, i.e.,
$$U\left(\varepsilon\right) = \sum_{i=1}^{n}\| D^{SVM}_{Y^*(\varepsilon)}\left(x_i(\varepsilon)\right) - D^{SVM}_{Y^*}(x_i) \|_p$$
where $\|\cdot\|_p$ is $l_p$ norm ($1 \leq p \leq \infty$) or $\| \cdot \|_0$, which counts the nonzero elements of a vector. 

For clustering problems, recall the aim of clustering is to partition the data points into different groups, i.e., $D^{CVC}_{Y^*(\varepsilon)} $ is a partition of points in $X(\varepsilon)$, there are many different ways of measuring the clustering results. See \cite{yin2024rapid,feng2023review} for some of the clustering functions. 
One natural way is to use a  matrix to represent the partition of points. Take $\mathcal{V}= \{ V_1, \cdots, V_K \}$ as an example. A 0-1 matrix $\widehat{D}(X) \in \mathbb{R}^{n \times K}$ is defined as follows: 
\begin{equation*}
	\widehat{D}(X) _{ij}=\begin{cases}
		1,\ \ &\ \text{if}\ x_i\in V_j,\\
		0,\ \ &\ \text{otherwise}.
	\end{cases}
\end{equation*}
Then the matrix $ \widehat{D}(X) \widehat{D}(X)^T$ actually shows whether data points are grouped together, where
\begin{equation*}
	\left(\widehat{D}(X) \widehat{D}(X)^T\right)_{ij}=\begin{cases}
		1,\ \ &\ \text{if}\ x_i, x_j\ \text{are in the same partition},\\
		0,\ \ &\ \text{otherwise}.
	\end{cases}
\end{equation*}
The following function is proposed by Biggio et al.  \cite{biggio2013data} 
\begin{equation}\label{eq-U}
	\delta(\varepsilon) = \left\| \widehat{D}(Y^*(\varepsilon)) \widehat{D}(Y^*(\varepsilon))^T - \widehat{D}(Y^*) \widehat{D}(Y^*)^T \right\|_F^2.
\end{equation}
Chhabra et al.  \cite{chhabra2020suspicion} believed that $\delta$ increases with the number of points that spill over from partition $V_1$ to $V_2$ for $K=2$. However, it is still not quite clear about whether this function can fully represent the deviation of $D_{Y^*(\varepsilon)}$ over the original $D_{Y^*}$. Therefore, below we conduct a systematic analysis on the property of  $\delta$ defined in \eqref{eq-U}. We consider the following two scenarios. 

\subsection{Analysis on 2-Way Clustering}
$K=2$, with $X \in \mathbb{R}^{d \times n}$ clustered into $V_1$ and $V_2$ with $|V_1|=n_1,\ |V_2|=n_2$. Let $n = n_1 + n_2$. Assume that under perturbation $\varepsilon$, the clustering of $X(\varepsilon)$ is changed to $V_1 \setminus S$, $V_2 \cup S$, where $S \subseteq V_1$, $|S|=s$. We have the following result.

\begin{Theorem}\label{thm-2w}
	For 2-way clustering, let $\delta\left(\varepsilon\right) $ be defined by \eqref{eq-U}. 
	\begin{itemize}
		\item [(i)] $	\delta\left(\varepsilon\right) = 2s(n - s)$.
		\item [(ii)] If  $s<\min(n_1, \lceil \frac{n}{2} \rceil)$\footnote{Here $\lceil a \rceil$ denotes the smallest integer that is greater than or equal to $a$.},  $\delta\left(\varepsilon\right) $ is a deviation function.
	\end{itemize}
\end{Theorem}
\textbf{Proof}: For simplicity, we use $\widehat{D}(\varepsilon) $ and $\widehat{D}(0)$ to represent $\widehat{D}(Y^*(\varepsilon))$ and $\widehat{D}(Y^*)$. 
For (i), without loss of generality, let $V_1=\{x_1,\cdots,x_{n_1}\},\ V_2=\{x_{n_1+1},\cdots,x_{n}\}$. 
Then it holds that
\begin{equation*}
	\widehat{D}(0)=\begin{bmatrix}
		e_{n_1}&0\\
		0&e_{n_2}
	\end{bmatrix}\in \mathbb{R}^{n \times 2},\ \ \ 
	\widehat{D}(0)\widehat{D}(0)^T=\begin{bmatrix}
		E_{n_1}&0\\
		0&E_{n_2}
	\end{bmatrix}\in \mathbb{R}^{n \times n},
\end{equation*}
where $e_{n_1}$ is the column vector of length $n_1$ whose elements are all ones and $E_{n_1}$ denotes the matrix of size $n_1\times n_1$ whose elements are all ones. Here '0' denotes the zero vector or matrix of proper sizes.
By changing the last $s$ data points in $V_1$ to $V_2$, we have
\begin{equation*}
	\widehat{D}(\varepsilon)=\begin{bmatrix}
		e_{n_1-s}&0\\
		0&e_{n_2+s}
	\end{bmatrix}\in \mathbb{R}^{n \times 2},\ \ \ 
	\widehat{D}(\varepsilon) \widehat{D}(\varepsilon)^T=\begin{bmatrix}
		E_{n_1-s}&0\\
		0&E_{n_2-s}
	\end{bmatrix}\in \mathbb{R}^{n \times n},
\end{equation*}
leading to the following ($E_{i\times j}$ denotes the matrix of $i$ by $j$ with all elements one)
\begin{equation*}
	\widehat{D}(0)\widehat{D}(0)^T-\widehat{D}(\varepsilon) \widehat{D}(\varepsilon)^T=\begin{bmatrix}
		0&E_{(n_1-s)\times s}&0\\
		E_{s\times(n_1-s)}&0&-E_{s\times n_2}\\
		0&-E_{n_2\times s}&0
	\end{bmatrix}
\end{equation*}
and 
\begin{equation*}
	\delta\left(\varepsilon\right)  =\|\widehat{D}(0)\widehat{D}(0)^T-\widehat{D}(\varepsilon) \widehat{D}(\varepsilon)^T\|_F^2=2(n-s)s.
\end{equation*}
This gives (i).

To show (ii), obviously, $	\delta(0)  = 0$ and $	\delta\left(\varepsilon\right)   > 0$ for any $Y^*(\varepsilon) \neq Y^*(0)$. 
Moreover, note that $0 < s < n_1$, therefore, as $s$ increases for $s \in (0, \min(n_1, \lceil \frac{n}{2} \rceil))$,  $\delta(\cdot)$ is a nondecreasing function with respect to $s$. 
In other words, only when $s \in (0, \min(n_1, \lceil \frac{n}{2} \rceil))$, $\delta$ is a deviation function with respect to $s$. 
\hfill$\square$

We give some examples as follows. 

\begin{Example}\label{eg-2w-1}
	Let $X=[x_1,\cdots,x_5]$ with partition $V_1=\{x_1,\ x_2,\ x_3,\ x_4\}$ and $V_2=\{x_5\}$, which means $n_1=4$ and $n_2=1$. It holds that
	\begin{equation*}
		\widehat{D}(0)=\begin{bmatrix}
			e_{4}&0\\
			0&e_{1}
		\end{bmatrix}\in \mathbb{R}^{5 \times 2}.
	\end{equation*}
	Changing $x_4$ from $V_1$ to $V_2$, we get $V_1^{\prime}=\{x_1,\ x_2,\ x_3\}$ and $V_2^{\prime}=\{x_4,\ x_5\}$, leading to the following
	\begin{equation*}
		\widehat{D}(\varepsilon) =\begin{bmatrix}
			e_{3}&0\\
			0&e_{2}
		\end{bmatrix}.
	\end{equation*}
	Therefore, 	$\delta\left(\varepsilon\right) =\|	\widehat{D}(0)\widehat{D}(0)^T-\widehat{D}(\varepsilon) \widehat{D}(\varepsilon)^T\|_F^2=8.$
\end{Example}

\begin{Example}\label{eg-2w-2}
	Let $X$ and $V_1,\ V_2$ be the same as in Example \ref{eg-2w-1}. 
	Changing $x_3,\ x_4$ from $V_1$ to $V_2$,\ to get $V_1^{\prime}=\{x_1,\ x_2\}$ and $V_2^{\prime}=\{x_3,\ x_4,\ x_5\}$, we get 
	\begin{equation*}
		\widehat{D}(\varepsilon)=\begin{bmatrix}
			e_{2}&0\\
			0&e_{3}
		\end{bmatrix},
	\end{equation*}
	which gives $\delta\left(\varepsilon\right) =12$.
\end{Example}

\begin{Example}\label{eg-2w-3}
	Let $X$ and $V_1,\ V_2$ be the same as in Example \ref{eg-2w-1}. 
	Changing $x_2,\ x_3,\ x_4$ from $V_1$ to $V_2$ to get  $V_1^{\prime}=\{x_1\}$ and $V_2^{\prime}=\{x_2,\ x_3,\ x_4,\ x_5\}$, we get 
	\begin{equation*}
		\widehat{D}(\varepsilon)=\begin{bmatrix}
			e_{1}&0\\
			0&e_{4}
		\end{bmatrix},
	\end{equation*}
	which also leads to $\delta\left(\varepsilon\right) =12$.
\end{Example}

Comparing Example \ref{eg-2w-2} and Example \ref{eg-2w-3}, the number of changed data is increasing. However, the deviation function $\delta$ is the same. In this case, $s =3 > \lceil \frac{n}{2} \rceil$ in Example \ref{eg-2w-3}, implying that $\delta$ can not fully represent the deviation of the perturbed clustering results over the original clustering results. 

In summary, Theorem \ref{thm-2w} as well as the above examples show that choosing a proper deviation function $U$ is very important. However, sometimes it is difficult and even tricky to choose a good function which satisfy deviation properties (i)-(iii). Also, when dealing with adversarial learning models, we have to be very careful in order to choose a good deviation function since the chosen function may not fully reflect the changes in perturbation decision function after perturbation. 

\subsection{Analysis on 3-Way Clustering}
$K=3$, with $X \in \mathbb{R}^{d \times n}$ clustered into $V_1,\  V_2,\ V_3$ with $|V_i| = n_i,\ i=1,2,3$, and $n = \sum_{i=1}^3 n_i$. After perturbation, the cluster changes to $V_1 \setminus (S_1 \cup S_2)$, $V_2 \cup S_1$, $V_3 \cup S_2$, where $|S_1|=s_1$, $|S_2|=s_2$. 
\begin{Theorem}\label{thm-3w}
	For 3-way clustering, it holds that
	\begin{equation*}
		\delta\left(\varepsilon\right)  = (s_1+s_2)\left(2n_1-(s_1+s_2)\right)+s_1(2n_2-s_1)+s_2(2n_3-s_2).
	\end{equation*}
	In particular, 
	\begin{itemize}
		\item[(i)] If $S_2 = \emptyset$,\ $\delta\left(\varepsilon\right)  =2s_1(n-s_1) $.\ Then $\delta(\cdot)$ is a deviation function.
		\item [(ii)] If $n_1 = n_2 = n_3$ and $s_1 = s_2$,  $\delta\left(\varepsilon\right)  =2s_1\left(\frac{4}{3}n - 3s_1\right)$. Then $\delta(\cdot)$ is a deviation function.
		\item [(iii)] If $s_1 = s_2$,  $\delta\left(\varepsilon\right) =s_1(4n_1+2n_2+2n_3 - 6s_1)$. For $s_1 \in \left(0, \min\left( \lceil\frac{2n_1 + n_2 + n_3}{6} \rceil, \lceil\frac{n_1}{2} \rceil\right)\right)$, $\delta(\cdot)$ is a deviation function.		
	\end{itemize}	
\end{Theorem}
\textbf{Proof.} Without loss of generality, assume that $V_1=\{x_1,\cdots,x_{n_1}\}$,  $V_2=\{x_{n_1+1},\cdots,x_{n_1+n_2}\}$,  $V_3=\{x_{n_1+n_2+1},\cdots,x_{n}\}$. 
It holds that
\begin{equation*}
	\widehat{D}(0)=\begin{bmatrix}
		e_{n_1}&0&0\\
		0&e_{n_2}&0\\
		0&0& e_{n_3}
	\end{bmatrix}\in \mathbb{R}^{n \times 3},\ \ \ 
	\widehat{D}(0)\widehat{D}(0)^T=\begin{bmatrix}
		E_{n_1}&0&0\\
		0&E_{n_2}&0\\
		0&0&E_{n_3}
	\end{bmatrix}\in \mathbb{R}^{n \times n}.
\end{equation*}

Now a subset of data points $S_1 \subseteq V_1$ changes their cluster membership from $V_1$ to $V_2$ and a subset of data points $S_2 \subseteq V_1$ changes their cluster membership from $V_1$ to $V_3$. It holds that
\footnotesize
\begin{equation*}
	\widehat{D}(\varepsilon) =\begin{bmatrix}
		e_{n_1-(s_1+s_2)}&0&0\\
		0&e_{s_1}&0\\
		0&0& e_{s_2}\\
		0&e_{n_2}&0\\
		0&0& e_{n_3}\\
	\end{bmatrix}\in \mathbb{R}^{n \times 3},\ \ 
	\widehat{D}(\varepsilon) \widehat{D}(\varepsilon)^T=
	\begin{bmatrix}
		E_{n_1-(s_1+s_2)}&0&0&0&0\\		
		0&E_{s_1}&0&E_{s_1\times n_2}&0\\		
		0&0&E_{s_2}&0&E_{s_2\times n_3}\\	
		0&E_{n_2\times s_1}&0&E_{n_2}&0\\		
		0&0&E_{n_3\times s_2}&0&E_{n_3}\\
	\end{bmatrix}\in \mathbb{R}^{n \times n},
\end{equation*}
leading to the following
\begin{equation*}
	\hat{D}(0)\hat{D}(0)^T-\hat{D}(\varepsilon) \hat{D}(\varepsilon)^T
	=\begin{bmatrix}
		0&E_{(n_1-(s_1+s_2))\times s_1}&E_{(n_1-(s_1+s_2))\times s_2}&0&0\\		
		E_{s_1\times (n_1-(s_1+s_2))}&0&E_{s_1\times s_2}&-E_{s_1\times n_2}&0\\		
		E_{s_2\times (n_1-(s_1+s_2))}&E_{s_2\times s_1}&0&0&-E_{s_2\times n_3}\\		
		0&-E_{n_2\times s_1}&0&0&0\\		
		0&0&-E_{n_3\times s_2}&0&0\\
	\end{bmatrix}.
\end{equation*}
\normalsize 
After calculation, we get
\begin{equation*}
	\begin{split}
		\delta(\varepsilon) &=\|\widehat{D}(0)\widehat{D}(0)^T-\widehat{D}(\varepsilon) \widehat{D}(\varepsilon)^T\|_F^2\\
		&=(n_1-(s_1+s_2))(s_1+s_2)+s_1(n_1-s_1+n_2)+s_2(n_1-s_2+n_3)+n_2s_1+n_3s_2\\
		&=2(n_1(s_1+s_2)+n_2s_1+n_3s_2)-((s_1+s_2)^2+s_1^2+s_2^2)\\
		&=(s_1+s_2)(2n_1-(s_1+s_2))+s_1(2n_2-s_1)+s_2(2n_3-s_2).
	\end{split}
\end{equation*}  
This gives the first part of the results.

For (i), $S_2 = \emptyset$. In this case, the third cluster $V_3$ does not play any role. Then $\delta(\cdot)$ reduces to $2s_1(n_1 + n_2 - s_1)$,  which coincides with the results in Theorem \ref{thm-2w}. 

For (ii),\ if $n_1 = n_2 = n_3$,\ $s_1 = s_2$,\ $\delta(\cdot)$ takes the following form (note that $s=s_1+s_2=2s_1$)
\begin{equation*}
	\delta (\varepsilon)
	= 2s_1(4n_1 - 3s_1)
	=2s_1\left(\frac{4}{3}n - 3s_1\right).
\end{equation*}
Note that for $s_1 \in \left(0, \lceil \frac{2n}{9} \rceil\right),\ \delta$ is nondecreasing. 
Also note that $s_1<\lceil\frac{n_1}{2}\rceil=\lceil\frac{n}{6}\rceil$, therefore, for any $s_1$ in that case, $s_1<\lceil \frac{2n}{9} \rceil$, which means that $\delta(\cdot)$ is always nondecreasing. That is, for any $s_1 $, $\delta(\cdot)$ is always a deviation function.

For (iii), if $s_1 = s_2$, then 
\begin{equation*}
	\begin{split}
		\delta\left(\varepsilon\right) &= 2s_1(2n_1 - 2s_1) + s_1(2n_2 - s_1) + s_1(2n_3 - s_1) \\
		&= s_1(4n_1+2n_2+2n_3 - 6s_1) . \\
	\end{split}
\end{equation*}
So for $s_1 \in \left(0, \min\left( \lceil\frac{2n_1 + n_2 + n_3}{6} \rceil, \lceil\frac{n_1}{2} \rceil\right)\right),\ \delta(\cdot)$ is nondecreasing. Therefore, it is a deviation function.
The proof is finished.
\hfill$\square$

Below we show some examples (Example \ref {eg-3w-c2-1} and  \ref {eg-3w-c2-2}) for case (ii) and case (iii) (Example \ref {eg-3w-c3-1}-\ref {eg-3w-c3-3}).
\begin{Example}\label{eg-3w-c2-1}
	Let $X=[x_1,\cdots,x_{15}]$ with partition $V_1=\{x_1,\cdots,x_5\}$,  $V_2=\{x_6,\cdots,x_{10}\}$ and $V_3=\{x_{11},\cdots,x_{15}\}$, that is, $n_1=n_2=n_3=5$. It holds that
	\begin{equation*}
		\widehat{D}(0)=\begin{bmatrix}
			e_{5}&0&0\\
			0&e_{5}&0\\
			0&0& e_{5}
		\end{bmatrix}\in \mathbb{R}^{15 \times 3},\ \ \ 
		\widehat{D}(0)\widehat{D}(0)^T=\begin{bmatrix}
			E_{5}&0&0\\
			0&E_{5}&0\\
			0&0&E_{5}
		\end{bmatrix}.
	\end{equation*}     
	Changing $x_2$ from $V_1$ to $V_2$ and $x_3$ from $V_1$ to $V_3$, which means $s_1=s_2=1$,  we get
	\begin{equation*}
		\widehat{D}(\varepsilon) =\begin{bmatrix}
			e_{3}&0&0\\
			0&e_{1}&0\\
			0&0& e_{1}\\
			0&e_{5}&0\\
			0&0& e_{5}\\
		\end{bmatrix},\ \ \ 
		\widehat{D}(\varepsilon) \widehat{D}(\varepsilon)^T=\begin{bmatrix}
			E_{3}&0&0&0&0\\
			
			0&1&0&E_{1\times 5}&0\\
			
			0&0&1&0&E_{1\times 5}\\
			
			0&E_{5\times 1}&0&E_{5}&0\\
			
			0&0&E_{5\times 1}&0&E_{5}\\
		\end{bmatrix}.
	\end{equation*}
	It holds that
	\begin{equation*}
		\widehat{D}(0)\widehat{D}(0)^T-\widehat{D}(\varepsilon) \widehat{D}(\varepsilon)^T=\begin{bmatrix}
			0&E_{3\times1}&E_{3\times1}&0&0\\
			
			E_{1\times3}&0&1&-E_{1\times 5}&0\\
			
			E_{1\times3}&1&0&0&-E_{1\times 5}\\
			
			0&-E_{5\times 1}&0&0&0\\
			
			0&0&-E_{5\times 1}&0&0\\
		\end{bmatrix},
	\end{equation*} 
	and $	\delta\left(\varepsilon\right)=\|\widehat{D}(0)\widehat{D}(0)^T-\widehat{D}(\varepsilon) \widehat{D}(\varepsilon)^T\|_F^2=34.$
\end{Example}

\begin{Example}\label{eg-3w-c2-2}
	Let $X$ and $V_1,\ V_2,\ V_3$ be the same as in Example \ref{eg-3w-c2-1}. 	$\widehat{D}(0)$ is the same as in Example \ref{eg-3w-c2-1}.  
	Changing $x_2,\ x_3$ from $V_1$ to $V_2$ and $x_4,\ x_5$ from $V_1$ to $V_3$, which means $s_1=s_2=2$,  we get
	\begin{equation*}
		\widehat{D}(\varepsilon) =\begin{bmatrix}
			e_{1}&0&0\\
			0&e_{2}&0\\
			0&0& e_{2}\\
			0&e_{5}&0\\
			0&0& e_{5}\\
		\end{bmatrix},\ \ \ 
		\widehat{D}(\varepsilon) \widehat{D}(\varepsilon)^T=\begin{bmatrix}
			1&0&0&0&0\\
			
			0&E_2&0&E_{2\times 5}&0\\
			
			0&0&E_2&0&E_{2\times 5}\\
			
			0&E_{5\times 2}&0&E_{5}&0\\
			
			0&0&E_{5\times 2}&0&E_{5}\\
		\end{bmatrix}.
	\end{equation*}
	It holds that
	\begin{equation*}
		\widehat{D}(0)\widehat{D}(0)^T-\widehat{D}(\varepsilon) \widehat{D}(\varepsilon)^T=\begin{bmatrix}
			0&E_{1\times2}&E_{1\times2}&0&0\\
			
			E_{2\times1}&0&E_2&-E_{2\times 5}&0\\
			
			E_{2\times1}&E_2&0&0&-E_{2\times 5}\\
			
			0&-E_{5\times 2}&0&0&0\\
			
			0&0&-E_{5\times 2}&0&0\\
		\end{bmatrix},
	\end{equation*} 
	and $	\delta\left(\varepsilon\right)=\|\widehat{D}(0)\widehat{D}(0)^T-\widehat{D}(\varepsilon) \widehat{D}(\varepsilon)^T\|_F^2=56.$
\end{Example}
For Example \ref {eg-3w-c2-1} and  Example \ref {eg-3w-c2-2}, it can be noticed that for case (ii), if $s_1$ increases, then $\delta(\cdot)$ increases, which fully reflect the changes of $\varepsilon$ in $\delta$-measure function.

\begin{Example}\label{eg-3w-c3-1}
	Let $X=[x_1,\cdots,x_{11}]$ with partition  $V_1=\{x_1,\cdots,x_9\}$,  $V_2=\{x_{10}\}$ and $V_3=\{x_{11}\}$, i.e., $n_1=9$, $n_2=1$ and $n_3=1$. It holds that
	\begin{equation*}
		\widehat{D}(0)=\begin{bmatrix}
			e_{9}&0&0\\
			0&e_{1}&0\\
			0&0& e_{1}
		\end{bmatrix}\in \mathbb{R}^{11 \times 3},\ \ \ 
		\widehat{D}(0)\widehat{D}(0)^T=\begin{bmatrix}
			E_{9}&0&0\\
			0&1&0\\
			0&0&1
		\end{bmatrix}.
	\end{equation*}     
	Changing $x_6,\ x_7$ from $V_1$ to $V_2$ and $x_8,\ x_9$ from $V_1$ to $V_3$, which means $s_1=s_2=2$,  we get
	\[
	\widehat{D}(\varepsilon)=
	\begin{bmatrix}
		e_{5}&0&0\\
		0&e_{2}&0\\
		0&0&e_{2}\\
		0&e_{1}&0\\
		0&0&e_{1}
	\end{bmatrix},\ \ \ 
	\widehat{D}(\varepsilon)\widehat{D}(\varepsilon)^{T}=
	\begin{bmatrix}
		E_{5}&0&0&0&0\\
		0&E_{2}&0&E_{2\times 1}&0\\
		0&0&E_{2}&0&E_{2\times 1}\\
		0&E_{1\times 2}&0&E_{1}&0\\
		0&0&E_{1\times 2}&0&E_{1}
	\end{bmatrix}.
	\]
	It holds that
	\[
	\widehat{D}(0)\widehat{D}(0)^{T}-\widehat{D}(\varepsilon)\widehat{D}(\varepsilon)^{T}=
	\begin{bmatrix}
		0&E_{5\times 2}&E_{5\times 2}&0&0\\
		E_{2\times 5}&0&E_{2}&-E_{2\times 1}&0\\
		E_{2\times 5}&E_{2}&0&0&-E_{2\times 1}\\
		0&-E_{1\times 2}&0&0&0\\
		0&0&-E_{1\times 2}&0&0
	\end{bmatrix},
	\]
	and $$\delta\left(\varepsilon\right)=\|\widehat{D}(0)\widehat{D}(0)^T-\widehat{D}(\varepsilon) \widehat{D}(\varepsilon)^T\|_F^2=56.$$
\end{Example}

\begin{Example}\label{eg-3w-c3-2}
	Let $X$ and $V_1,\ V_2,\ V_3$ be the same as in Example \ref{eg-3w-c3-1}. 	$\widehat{D}(0)$ is the same as in Example \ref{eg-3w-c3-1}.  
	Changing $x_4,\ x_5,\ x_6$ from $V_1$ to $V_2$ and $x_7,\ x_8,\ x_9$ from $V_1$ to $V_3$, which means $s_1=s_2=3$,  we get
	\[
	\widehat{D}(\varepsilon)=
	\begin{bmatrix}
		e_{3}&0&0\\
		0&e_{3}&0\\
		0&0&e_{3}\\
		0&e_{1}&0\\
		0&0&e_{1}
	\end{bmatrix},\qquad
	\widehat{D}(\varepsilon)\widehat{D}(\varepsilon)^{T}=
	\begin{bmatrix}
		E_{3}&0&0&0&0\\
		0&E_{3}&0&E_{3\times 1}&0\\
		0&0&E_{3}&0&E_{3\times 1}\\
		0&E_{1\times 3}&0&E_{1}&0\\
		0&0&E_{1\times 3}&0&E_{1}
	\end{bmatrix}.
	\]
	It holds that
	\[
	\widehat{D}(0)\widehat{D}(0)^{T}-\widehat{D}(\varepsilon)\widehat{D}(\varepsilon)^{T}=
	\begin{bmatrix}
		0&E_{3\times 3}&E_{3\times 3}&0&0\\
		E_{3\times 3}&0&E_{3}&-E_{3\times 1}&0\\
		E_{3\times 3}&E_{3}&0&0&-E_{3\times 1}\\
		0&-E_{1\times 3}&0&0&0\\
		0&0&-E_{1\times 3}&0&0
	\end{bmatrix},
	\]
	and $$\delta\left(\varepsilon\right)=\|\widehat{D}(0)\widehat{D}(0)^T-\widehat{D}(\varepsilon) \widehat{D}(\varepsilon)^T\|_F^2=66.$$
\end{Example}

\begin{Example}\label{eg-3w-c3-3}
	Let $X$ and $V_1,\ V_2,\ V_3$ be the same as in Example \ref{eg-3w-c3-1}. 	$\widehat{D}(0)$ is the same as in Example \ref{eg-3w-c3-1}.  
	Changing $\{x_2,\cdots,x_5\}$ from $V_1$ to $V_2$ and $\{x_6,\cdots,x_9\}$  from $V_1$ to $V_3$, which means $s_1=s_2=4$,  we get
	\[
	\widehat{D}(\varepsilon)=
	\begin{bmatrix}
		e_{1}&0&0\\
		0&e_{4}&0\\
		0&0&e_{4}\\
		0&e_{1}&0\\
		0&0&e_{1}
	\end{bmatrix},\ \ \ 
	\widehat{D}(\varepsilon)\widehat{D}(\varepsilon)^{T}=
	\begin{bmatrix}
		E_{1}&0&0&0&0\\
		0&E_{4}&0&E_{4\times 1}&0\\
		0&0&E_{4}&0&E_{4\times 1}\\
		0&E_{1\times 4}&0&E_{1}&0\\
		0&0&E_{1\times 4}&0&E_{1}
	\end{bmatrix}.
	\]
	It holds that
	\[
	\widehat{D}(0)\widehat{D}(0)^{T}-\widehat{D}(\varepsilon)\widehat{D}(\varepsilon)^{T}=
	\begin{bmatrix}
		0&E_{1\times 4}&E_{1\times 4}&0&0\\
		E_{4\times 1}&0&E_{4}&-E_{4\times 1}&0\\
		E_{4\times 1}&E_{4}&0&0&-E_{4\times 1}\\
		0&-E_{1\times 4}&0&0&0\\
		0&0&-E_{1\times 4}&0&0
	\end{bmatrix},
	\]
	and $$\delta\left(\varepsilon\right)=\|\widehat{D}(0)\widehat{D}(0)^T-\widehat{D}(\varepsilon) \widehat{D}(\varepsilon)^T\|_F^2=64.$$
\end{Example}

Comparing Example \ref{eg-3w-c3-1} and Example \ref{eg-3w-c3-2}, $s_1<\min\left( \lceil\frac{2n_1 + n_2 + n_3}{6} \rceil, \lceil\frac{n_1}{2} \rceil\right)=4$, so it increases as the number of changed data points grows. However, when comparing Example \ref{eg-3w-c3-2} and Example \ref{eg-3w-c3-3}, the deviation function $\delta(\cdot)$ decreases. In this case, $s_1=4\notin(0,4)$ in Example \ref{eg-3w-c3-3}, implying that  $\delta(\cdot)$ cannot fully represent the deviation of the perturbed clustering results from the original clustering results.

However, if $n_1,\ n_2,\ n_3$ are not the same or $s_1 \neq s_2$, the situation is more complicated to analyze.
Some small examples are given below, as shown in Example \ref {eg-3w-c4-1}, Example \ref {eg-3w-c4-2} and Example \ref {eg-3w-c4-3}.

\begin{Example}\label{eg-3w-c4-1}
	Let $X$ and $V_1,\ V_2,\ V_3$ be the same as in Example \ref{eg-3w-c3-1}. 	$\widehat{D}(0)$ is the same as in Example \ref{eg-3w-c3-1}.  
	Changing $x_4,\ x_5$ from $V_1$ to $V_2$ and $\{x_6,\cdots,x_9\}$ from $V_1$ to $V_3$, which means $s_1=2,\ s_2=4$,  we get
	\[
	\widehat{D}(\varepsilon)=
	\begin{bmatrix}
		e_{3}&0&0\\
		0&e_{2}&0\\
		0&0&e_{4}\\
		0&e_{1}&0\\
		0&0&e_{1}
	\end{bmatrix},\ \ \ 
	\widehat{D}(\varepsilon)\widehat{D}(\varepsilon)^T=
	\begin{bmatrix}
		E_{3}&0&0&0&0\\
		0&E_{2}&0&E_{2\times 1}&0\\
		0&0&E_{4}&0&E_{4\times 1}\\
		0&E_{1\times 2}&0&E_{1}&0\\
		0&0&E_{1\times 4}&0&E_{1}
	\end{bmatrix}.
	\]
	It holds that
	\[
	\widehat{D}(0)\widehat{D}(0)^T-\widehat{D}(\varepsilon)\widehat{D}(\varepsilon)^T=
	\begin{bmatrix}
		0&E_{3\times 2}&E_{3\times 4}&0&0\\
		E_{2\times 3}&0&E_{2}&-E_{2\times 1}&0\\
		E_{4\times 3}&E_{4}&0&0&-E_{4\times 1}\\
		0&-E_{1\times 2}&0&0&0\\
		0&0&-E_{1\times 4}&0&0
	\end{bmatrix},
	\]
	and $\delta\left(\varepsilon\right)=64.$
\end{Example}

\begin{Example}\label{eg-3w-c4-2}
	Let $X$ and $V_1,\ V_2,\ V_3$ be the same as in Example \ref{eg-3w-c3-1}. 	$\widehat{D}(0)$ is the same as in Example \ref{eg-3w-c3-1}.  
	Changing $x_3,\ x_4$\ from $V_1$ to $V_2$ and $\{x_5,\cdots,x_9\}$ from $V_1$ to $V_3$, which means $s_1=2,\ s_2=5$,  we get
	\[
	\widehat{D}(\varepsilon)=
	\begin{bmatrix}
		e_{2}&0&0\\
		0&e_{2}&0\\
		0&0&e_{5}\\
		0&e_{1}&0\\
		0&0&e_{1}
	\end{bmatrix},\ \ \ 
	\widehat{D}(\varepsilon)\widehat{D}(\varepsilon)^T=
	\begin{bmatrix}
		E_{2}&0&0&0&0\\
		0&E_{2}&0&E_{2\times 1}&0\\
		0&0&E_{5}&0&E_{5\times 1}\\
		0&E_{1\times 2}&0&E_{1}&0\\
		0&0&E_{1\times 5}&0&E_{1}
	\end{bmatrix}.
	\]
	It holds that
	\[
	\widehat{D}(0)\widehat{D}(0)^T-\widehat{D}(\varepsilon)\widehat{D}(\varepsilon)^T=
	\begin{bmatrix}
		0&E_{2\times 2}&E_{2\times 5}&0&0\\
		E_{2\times 2}&0&E_{2}&-E_{2\times 1}&0\\
		E_{5\times 2}&E_{5}&0&0&-E_{5\times 1}\\
		0&-E_{1\times 2}&0&0&0\\
		0&0&-E_{1\times 5}&0&0
	\end{bmatrix},
	\]
	and $\delta\left(\varepsilon\right)=62.$
\end{Example}

\begin{Example}\label{eg-3w-c4-3}
	Let $X$ and $V_1,\ V_2,\ V_3$ be the same as in Example \ref{eg-3w-c3-1}. 	$\widehat{D}(0)$ is the same as in Example \ref{eg-3w-c3-1}.  
	Changing $x_3,\ x_4,\ x_5$ from $V_1$ to $V_2$ and $\{x_6,\cdots,x_9\}$ from $V_1$ to $V_3$, which means $s_1=3,\ s_2=4$,  we get
	\[
	\widehat{D}(\varepsilon)=
	\begin{bmatrix}
		e_{2}&0&0\\
		0&e_{3}&0\\
		0&0&e_{4}\\
		0&e_{1}&0\\
		0&0&e_{1}
	\end{bmatrix},\qquad
	\widehat{D}(\varepsilon)\widehat{D}(\varepsilon)^T=
	\begin{bmatrix}
		E_{2}&0&0&0&0\\
		0&E_{3}&0&E_{3\times 1}&0\\
		0&0&E_{4}&0&E_{4\times 1}\\
		0&E_{1\times 3}&0&E_{1}&0\\
		0&0&E_{1\times 4}&0&E_{1}
	\end{bmatrix}.
	\]
	It holds that
	\[
	\widehat{D}(0)\widehat{D}(0)^T-\widehat{D}(\varepsilon)\widehat{D}(\varepsilon)^T=
	\begin{bmatrix}
		0&E_{2\times 3}&E_{2\times 4}&0&0\\
		E_{3\times 2}&0&E_{3}&-E_{3\times 1}&0\\
		E_{4\times 2}&E_{4}&0&0&-E_{4\times 1}\\
		0&-E_{1\times 3}&0&0&0\\
		0&0&-E_{1\times 4}&0&0
	\end{bmatrix},
	\]
	and $\delta\left(\varepsilon\right)=66.$
\end{Example}

Comparing Example \ref{eg-3w-c4-1} with Example \ref{eg-3w-c4-2}, we see that although more points are changed, $\delta(\cdot)$ decreases. It indicates that in this situation, $\delta$-measure function are not fully reflect the changes in partitioning. Moreover, comparing Example \ref{eg-3w-c4-2} with Example \ref{eg-3w-c4-3} shows that even with the same number of changed points ($s_1+s_2=7$),\ $\delta(\cdot)$ can differ because the points are reassigned to different clusters. This highlights that $\delta(\cdot)$ is influenced not only by how many points are perturbed, but also by how those perturbations are distributed across clusters.

Theorem \ref{thm-3w} shows that the $\delta(\cdot)$ under 3-way  depends on two factors: how many points leave $V_1$ in total and how unevenly they split between $V_2$ and $V_3$. In the balanced, symmetric case (Example \ref{eg-3w-c2-1} and Example \ref{eg-3w-c2-2}),\ $\delta(\cdot)$ is a deviation function. When cluster sizes or perturbed sizes are unbalanced, the effect is more complicated.
In summary, $\delta(\cdot)$ is affected by both the number of perturbed points and the asymmetry of the reassignment, and it is predictable under balanced, symmetric case.

\subsection{Analysis on K-Way Clustering}
In this subsection, we extend the formula of $\delta$ to the general K-way clustering. 
The setting of K-way clustering is as follows. 
Let the original K-way clustering be
$V_1,\dots,V_K$ with $\ |V_k| = n_k,\ \sum_{k=1}^K n_k = n$. We consider the following setting.
After perturbation, we still have $K$ clusters, but only the first cluster $V_1$ spills out to other clusters:
\[
V_1' := V_1 \setminus \Big(\bigcup_{\ell=2}^K S_\ell\Big),\qquad
V_\ell' := V_\ell \cup S_\ell,\quad \ell = 2,\dots,K.
\]
where
$$
S_\ell \subseteq V_1,\ \  |S_\ell| = s_\ell,\ \  \ell=2,\dots,K, \ \ \ 
S := \sum_{\ell=2}^K s_\ell.
$$
In a word, we move in total $S$ points out of $V_1$, with $s_2$ points moved to $V_2$, $\dots$, and $s_K$ points moved to $V_K$.
\begin{Theorem}\label{thm-kw}
	For $K$-way clustering, it holds that
	\begin{equation*}
	\delta(\varepsilon)
	=
	\Big(\sum_{\ell=2}^K s_\ell\Big)
	\Big(2n_1 - \sum_{\ell=2}^K s_\ell\Big)
	+
	\sum_{\ell=2}^K s_\ell (2n_\ell - s_\ell)
	=	S(2n_1 - S)
	+
	\sum_{\ell=2}^K s_\ell (2n_\ell - s_\ell).
	\end{equation*}
\end{Theorem}

\textbf{Proof.} 
	Without loss of generality, assume that the data points are ordered so that
\[
V_1 = \{x_1,\dots,x_{n_1}\},\quad
V_2 = \{x_{n_1+1},\dots,x_{n_1+n_2}\},\quad
\dots,\quad
V_K = \{x_{n_1+\cdots+n_{K-1}+1},\dots,x_n\}.
\]
Recall that $\widehat D(0)\in\{0,1\}^{n\times K}$ is the
cluster-indicator matrix before perturbation. Hence the matrix
$
A := \widehat D(0)\widehat D(0)^T \in \{0,1\}^{n\times n}
$
has entries
\[
A_{ij} =
\begin{cases}
	1, & \text{if $x_i$ and $x_j$ belong to the same original cluster},\\[0.2em]
	0, & \text{otherwise}.
\end{cases}
\]
Similarly, let
$
B := \widehat D(\varepsilon)\widehat D(\varepsilon)^T ,
$
so that $B_{ij} = 1$ if and only if $x_i$ and $x_j$ are in the same cluster
after perturbation, and $B_{ij} = 0$ otherwise.

By definition,
$
\delta(\varepsilon)
= \|A - B\|_F^2
= \sum_{i=1}^n \sum_{j=1}^n (A_{ij} - B_{ij})^2. 
$
Since $A_{ij},B_{ij}\in\{0,1\}$, each term $(A_{ij}-B_{ij})^2=1$ if
and only if $A_{ij}\neq B_{ij}$, and $(A_{ij}-B_{ij})^2=0$ otherwise.
Therefore, $\delta(\varepsilon)$ counts the number of ordered pairs
$(i,j)$ for which the  relation of
$(x_i,x_j)$ changes after perturbation.
That is, 
$
\delta(\varepsilon)
= 2 \times \text{(number of unordered pairs $\{i,j\}$ whose relation changes)}. 
$
It thus suffices to count, under the given perturbation pattern, how many unordered pairs $\{i,j\}$ change their relation. There are two cases.

\textbf{Case 1.} Consider the points in $V_1$. 
After perturbation, the set $V_1$ is split into
\[
V_1' = V_1 \setminus \Big(\bigcup_{\ell=2}^K S_\ell\Big)
\quad\text{and}\quad
S_\ell \subseteq V_1,\ \ell=2,\dots,K,
\]
with $|V_1'| = n_1 - S$ and $|S_\ell| = s_\ell$.
We count the pairs that were in the same original cluster $V_1$ but end
up in different clusters.
\begin{itemize}
	\item[(i)]
	For pairs consisting of one point in $V_1'$ and one point in $ \bigcup_{\ell=2}^K S_\ell$, 
	there are $(n_1 - S)S$ such unordered pairs.
	\item[(ii)]
	For pairs consisting of one point in $S_\ell$ and one point in $S_{\ell'}$ with
	$\ell\neq\ell'$, 
	there are $\sum_{2\le \ell<\ell'\le K} s_\ell s_{\ell'}$ such unordered
	pairs.
\end{itemize}
All these pairs were in the same cluster (cluster $1$) originally, but are in
different clusters after perturbation, so their cluster relation changes.

\textbf{Case 2.} For each $\ell=2,\dots,K$, points in $S_\ell$ move from $V_1$ to $V_\ell$.
Before perturbation, any pair consisting of one point in $S_\ell$ and one point in
$V_\ell$ belonged to different clusters (cluster $1$ and cluster $\ell$),
whereas after perturbation they both belong to $V_\ell'$ and thus become
in the same cluster.
There are $s_\ell n_\ell$ such unordered pairs for each $\ell$,
and in total
$
\sum_{\ell=2}^K s_\ell n_\ell
$
unordered pairs whose relation changes in this way.

Overall, the total number of unordered pairs
whose relation changes is
$
\bigl(n_1 - S\bigr)S
+
\sum_{2\le \ell<\ell'\le K} s_\ell s_{\ell'}
+
\sum_{\ell=2}^K s_\ell n_\ell . 
$
Hence
\begin{equation*}
	\delta(\varepsilon)
	= 2\Bigg[
	\bigl(n_1 - S\bigr)S
	+
	\sum_{2\le \ell<\ell'\le K} s_\ell s_{\ell'}
	+
	\sum_{\ell=2}^K s_\ell n_\ell
	\Bigg].
\end{equation*}
Note that
$
S^2 = \Big(\sum_{\ell=2}^K s_\ell\Big)^2
= \sum_{\ell=2}^K s_\ell^2 + 2\sum_{2\le \ell<\ell'\le K} s_\ell s_{\ell'},
$
which implies
$
\sum_{2\le \ell<\ell'\le K} s_\ell s_{\ell'}
=
\frac{1}{2}\Big( S^2 - \sum_{\ell=2}^K s_\ell^2 \Big).
$
Therefore,
\begin{equation*}
	\delta(\varepsilon)
	= 2\Big[
	(n_1 - S)S
	+
	\frac{1}{2}\Big( S^2 - \sum_{\ell=2}^K s_\ell^2 \Big)
	+
	\sum_{\ell=2}^K s_\ell n_\ell
	\Big] 
	= S(2n_1 - S)
	+ \sum_{\ell=2}^K s_\ell(2n_\ell - s_\ell),
\end{equation*}
which is exactly the claimed formula.
This completes the proof.
\hfill$\square$

\section{Numerical Results}
\label{sec:numerics-bilevel}
In this part, we conduct numerical experiments to verify the theoretical
results obtained in Sections~\ref{sec2}–\ref{sec5}. 
The numerical tests are conducted in \textsc{Matlab} R2023a on a MacBook Air
(13-inch, M3, 2024) running macOS Sonoma~14.6 with an Apple M3 chip and
16~GB of memory.
The datasets used in our experiments are from the UCI Machine Learning Repository, which can be downloaded from https://archive.ics.uci.edu/datasets.

We consider the adversarial learning of convex clustering model \eqref{eq-cvc} with $p=2$.  Given data matrix $X = [x_1,\ldots,x_n]\in\mathbb{R}^{d\times n}$, we solve
\eqref{eq-cvc} by Ssnal \cite{sun2021convex} with default settings and obtain a baseline solution
$Y^*$ and cluster labels
$
\mathrm{cluster\_id}_0(i)\in\{1,\ldots,K\},\ i=1,\ldots,n. 
$
	For perturbed data $
	X(\varepsilon) = [x_1(\varepsilon),\ldots,x_n(\varepsilon)]
	$, Ssnal is also applied to obtain $Y^*(\varepsilon)$ and perturbed
	labels $\mathrm{cluster\_id}_\varepsilon(i)$. The number of changed labels compared to $Y^*$ is denoted by $N_{\mathrm{chg}}$. That is, 
	$
	N_{\mathrm{chg}}(\varepsilon)
	:=
	\bigl|\{i:\ \mathrm{cluster\_id}_\varepsilon(i)\neq\mathrm{cluster\_id}_0(i)\}\bigr|.
	$

From \cite[Theorems~12,~Theorems~13]{sun2021convex}, we know that every sequence generated by Ssnal converges to the unique optimal solution of the convex clustering problem. In addition, the semismooth Newton iterations enjoy local superlinear (and often quadratic) convergence under mild regularity conditions. These results ensure that, in our bilevel formulation, each lower-level convex clustering problem can be solved efficiently and we can obtain the unique globally optimal solution of \eqref{eq-pvar} for each $\varepsilon$.

\subsection{ Robustness Verification of the Convex Clustering Model}
\label{subsec:robustness}
In this part, we conduct numerical tests to verify the robustness of the convex clustering model. 
That is, when the perturbation $\varepsilon$ is relatively
small, the clustering result returned by the convex clustering model remains the same.

\begin{figure}[H]
	\centering
	\begin{minipage}{0.48\linewidth}
		\centering
		\includegraphics[width=\linewidth]{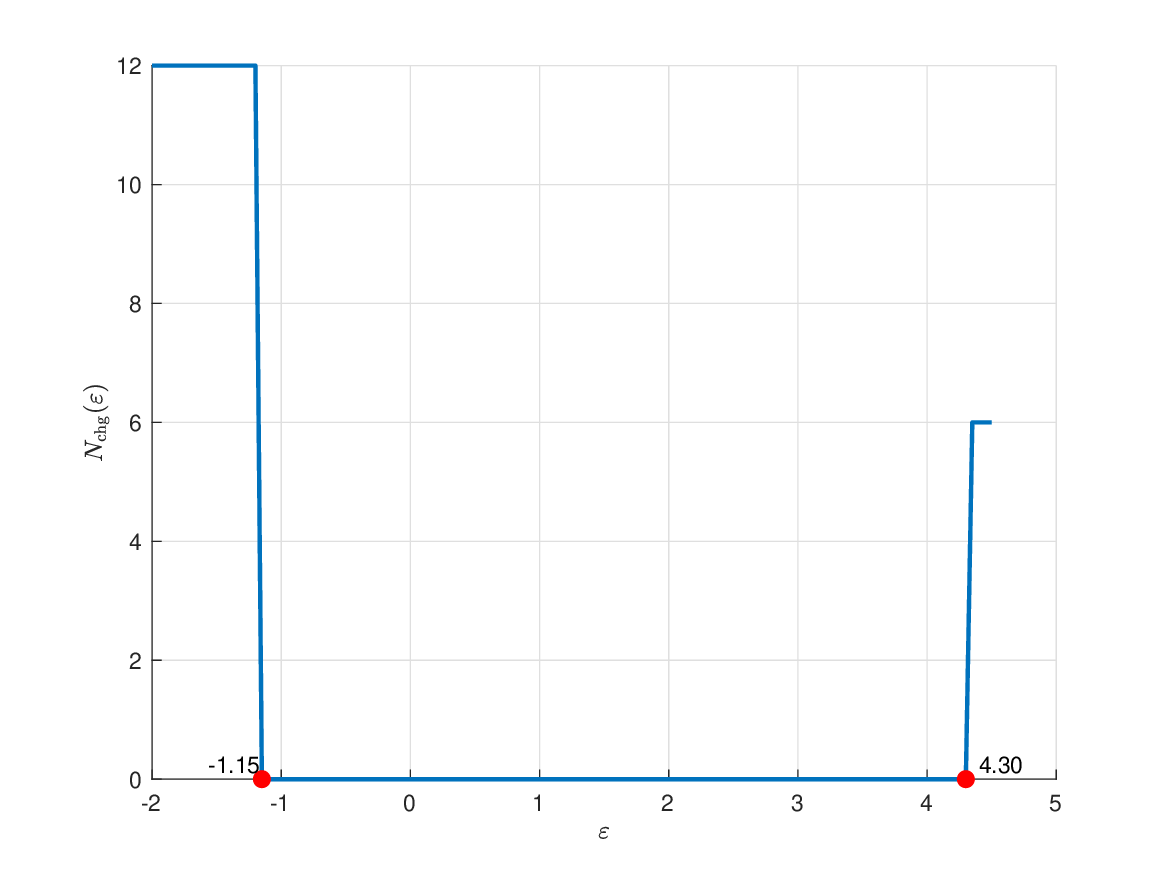}
	\end{minipage}\hfill
	\begin{minipage}{0.48\linewidth}
		\centering
		\includegraphics[width=\linewidth]{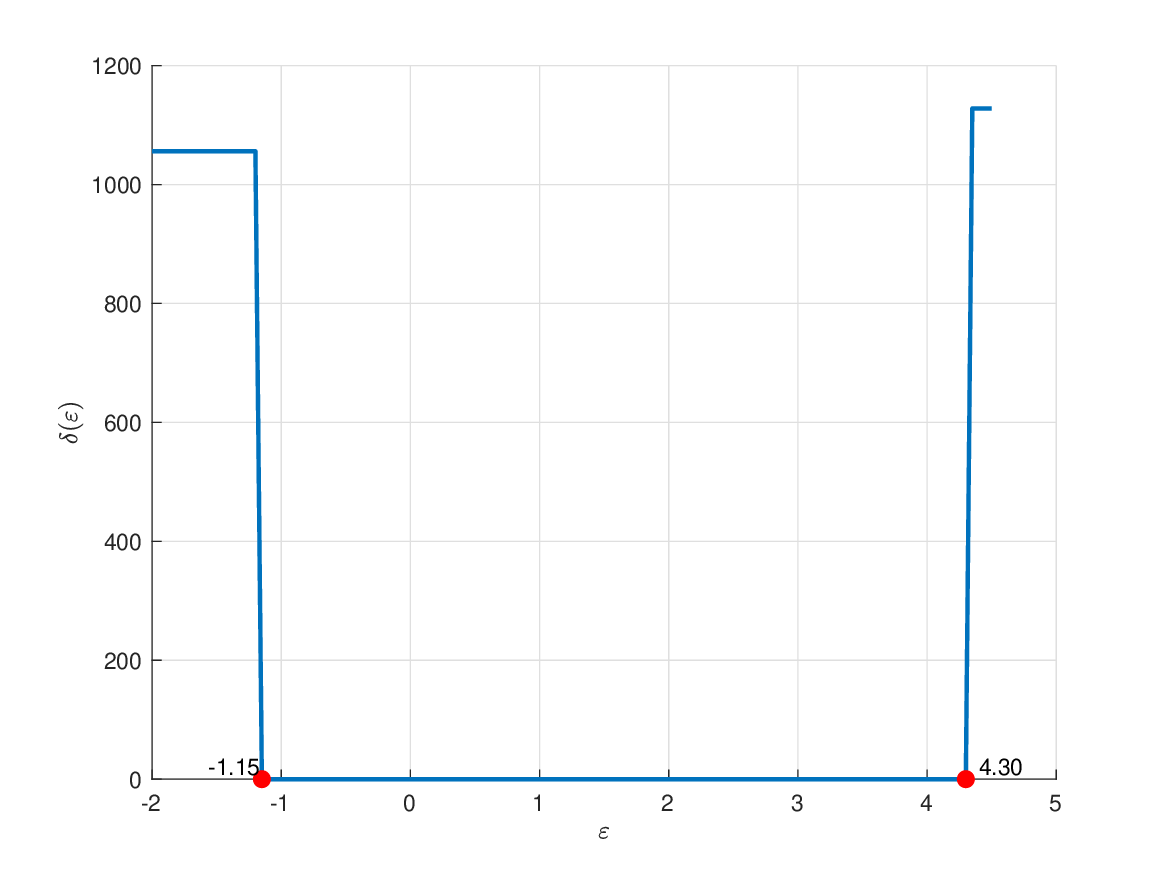}
	\end{minipage}
	\caption{
		The number of changed
		labels $N_{\mathrm{chg}}(\varepsilon)$ and deviation $\delta(\varepsilon)$  on Fisher Iris dataset.
		}
	\label{fig:iris-rb}
\end{figure}

	To better understand the role of perturbation, we select the Fisher Iris dataset with one feature (Feature 4). That is, $d=1$. 
	We choose additive noise as $x_i(\varepsilon) = x_i + \varepsilon$ for $i \in I_{\mathrm{pert}}$ and
	$x_i(\varepsilon) = x_i$ otherwise, where $I_{\mathrm{pert}} = \{1,2,3,4,5\}$. Other parameters are chosen as $n=100$, $\gamma = 500$. In this test, we increase the perturbation $|\varepsilon|$ gradually and report $N_{\mathrm{chg}}(\varepsilon)$ and $\delta(\varepsilon)$ in Figure~\ref{fig:iris-rb}.

		In Figure~\ref{fig:iris-rb}, it can be observed that when $\varepsilon\in (-1.15, 4.30)$, the number of changed labels is always 0, meaning that for $\varepsilon\in (-1.15, 4.30)$, the clustering results remains the same as the unperturbed result, which verify the robustness of the convex clustering model. 
		If $\varepsilon\geq4.30$, then  $N_{\mathrm{chg}}(\varepsilon)$ changes from 0 to 6, implying that there are six data whose label is changed, leading to the deviation function $\delta(\varepsilon)$ jump to 1128. In other words, if $\varepsilon>4.30$, attack happens. Similar situation also happens when $\varepsilon\leq-1.15$.

\subsection{Numerical Results on Deviation Functions}
\label{subsec:delta-discussion}
In this part, we consider adversarial learning by adding larger noise to the dataset. We will check by numerical results to see whether the $\delta(\cdot)$ is a reasonable deviation function.

\begin{figure}[H]
	\centering
	\begin{minipage}{0.48\linewidth}
		\centering
		\includegraphics[width=\linewidth]{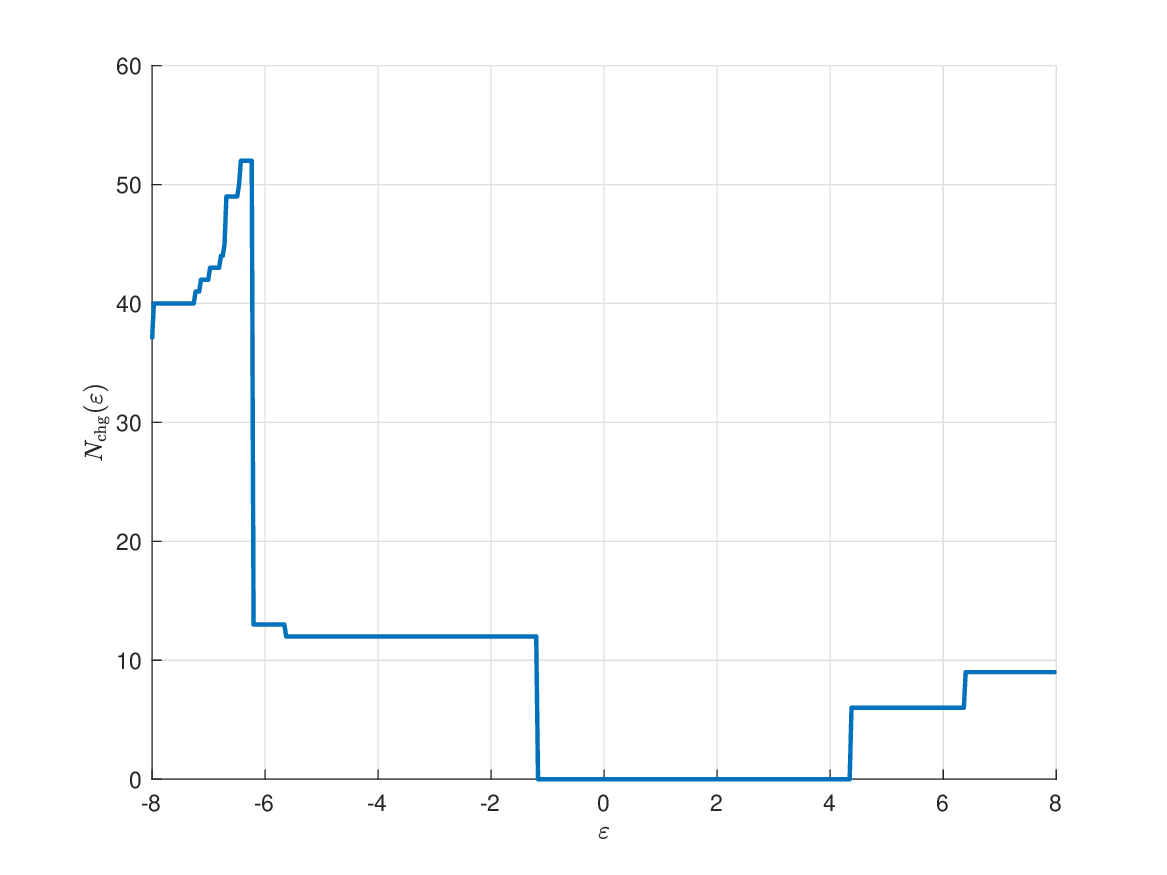}
	\end{minipage}\hfill
	\begin{minipage}{0.48\linewidth}
		\centering
		\includegraphics[width=\linewidth]{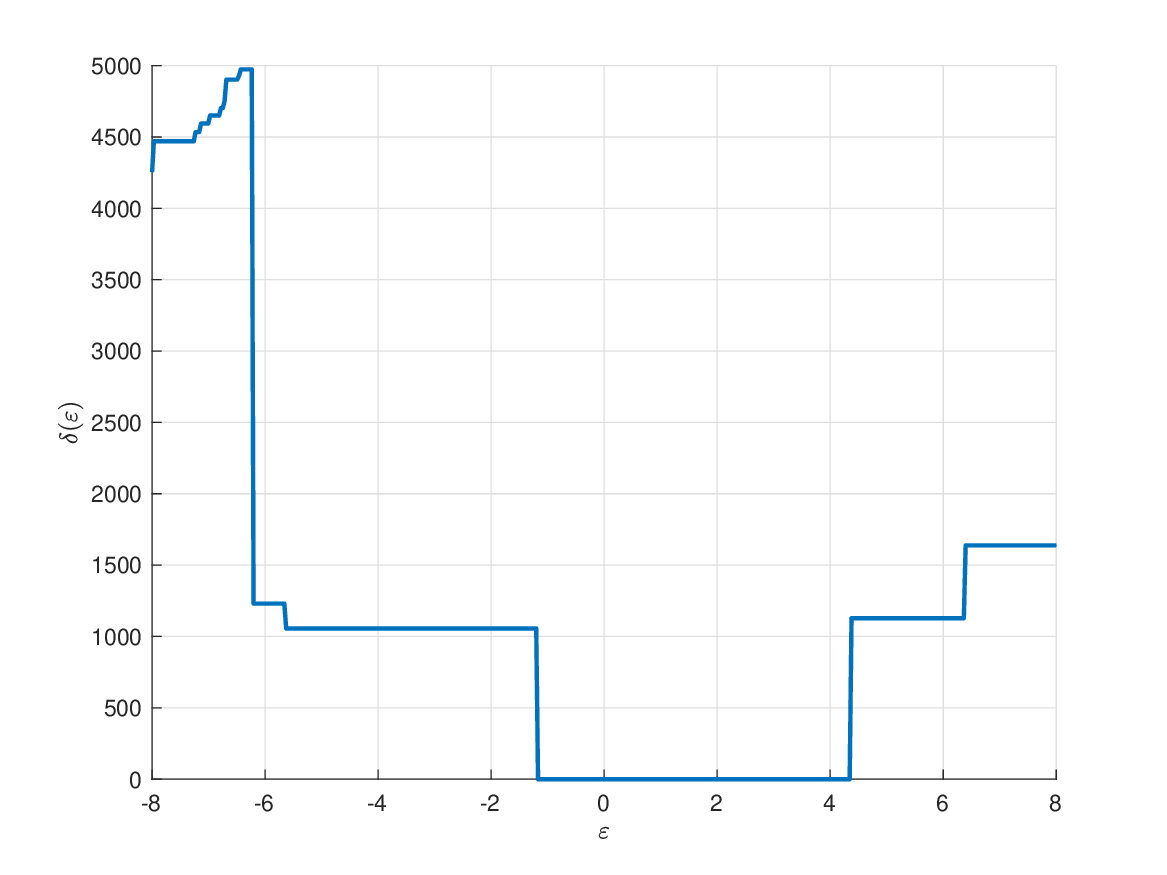}
	\end{minipage}
	\caption{
		The number of changed
		labels $N_{\mathrm{chg}}(\varepsilon)$ and deviation $\delta(\varepsilon)$ on Fisher Iris dataset. 
		}
	\label{fig:iris-edelta-es}
\end{figure}

The dataset and related parameters are the same as Section~\ref{subsec:robustness}. The results are demonstrated in Figure~\ref{fig:iris-edelta-es}. It can be noticed that as $|\varepsilon|$ increases, there are some points of $\varepsilon$ where $N_{\mathrm{chg}}$ jumps to another value. More interestingly, when $\varepsilon$ decreases from $-6$ to $-8$, the number of changed labels reaches its maximum value,  and then drops to $40$. Correspondingly,  the $\delta(\varepsilon)$ exhibits a very similar staircase structure, where jumps in $\delta(\varepsilon)$ coincide with jumps in
$N_{\mathrm{chg}}(\varepsilon)$, which verifies that the $\delta$-measure is a nondecreasing function with respect to $N_{\mathrm{chg}}$. 

\subsection{Verification of Bilevel Models}
\label{subsec:verify-bilevel}
\begin{figure}[H]
	\centering
	\begin{minipage}{0.48\linewidth}
		\centering
		\includegraphics[width=\linewidth]{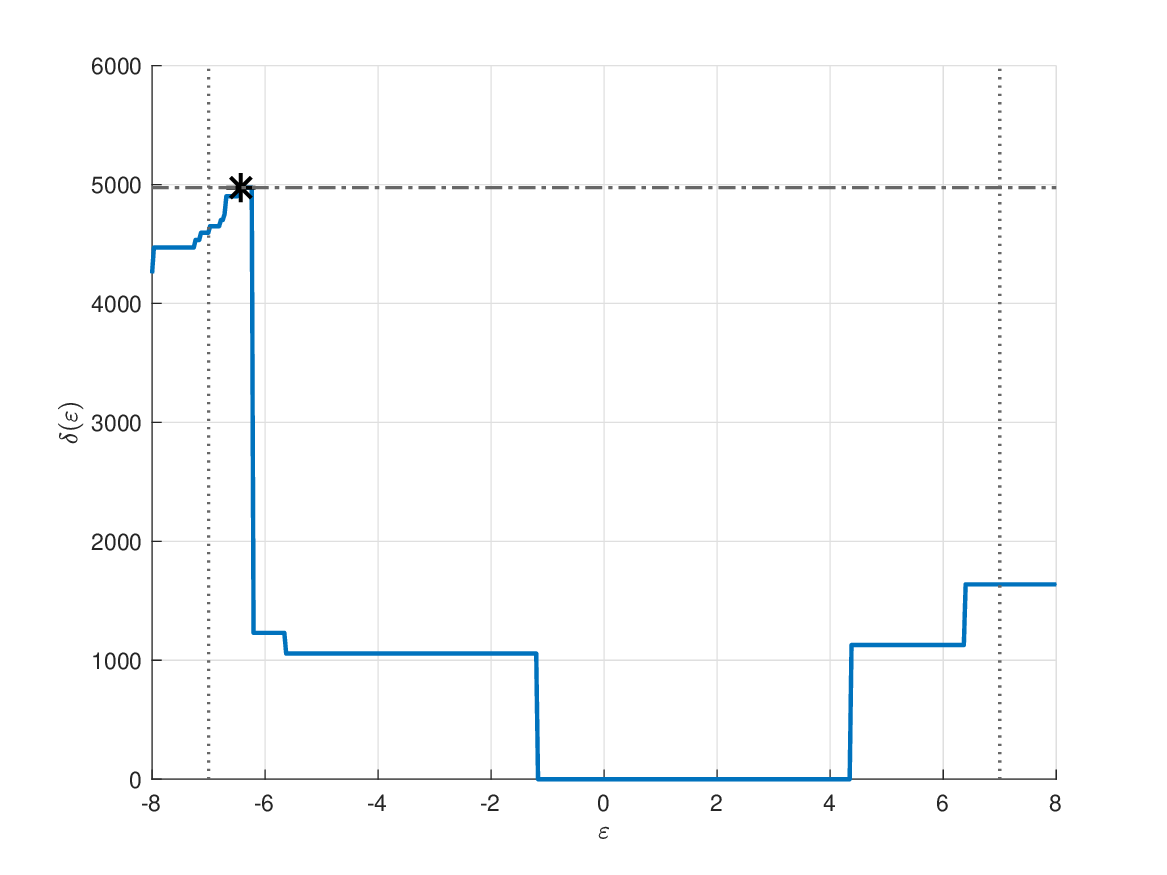}
	\end{minipage}\hfill
	\begin{minipage}{0.48\linewidth}
		\centering
		\includegraphics[width=\linewidth]{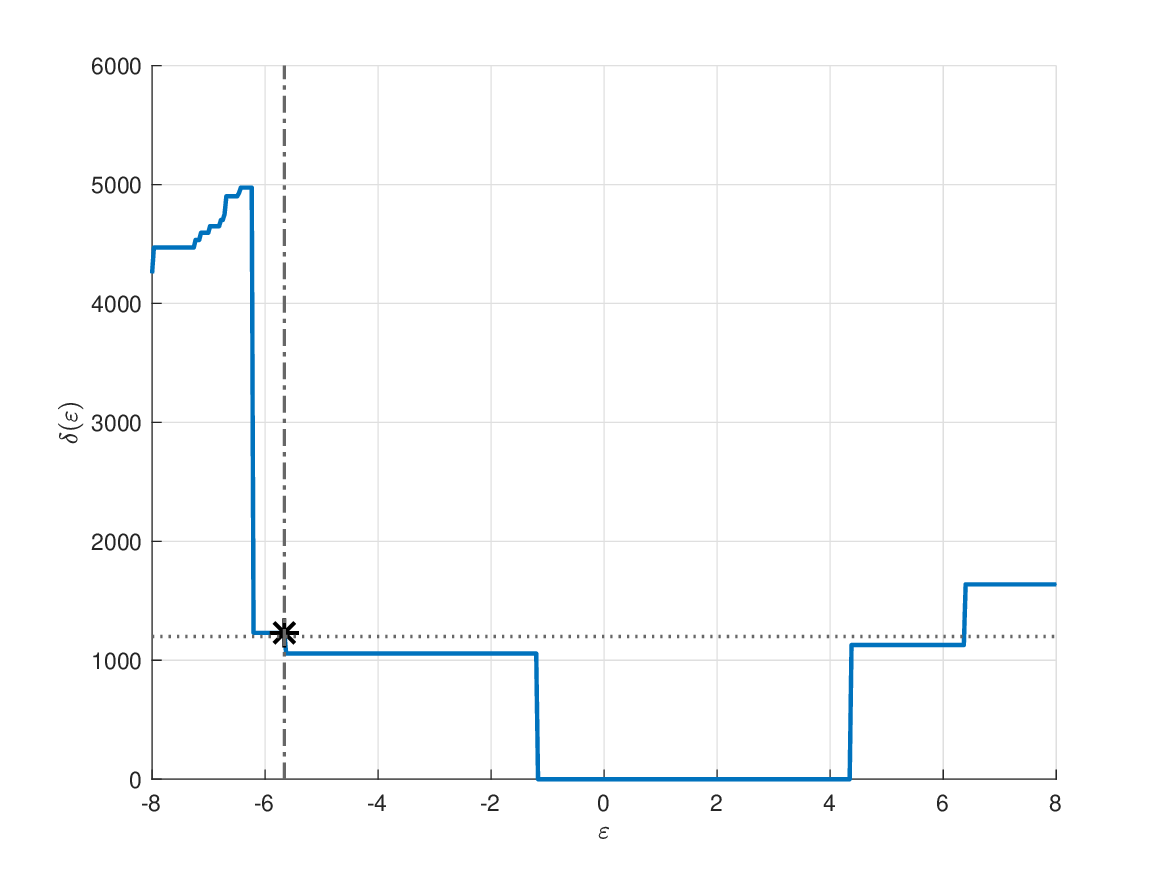}
	\end{minipage}
	\caption{
		Numerical illustration of the bilevel attack models \eqref{eq-adv1_cvc} and \eqref{eq-adv2_cvc} on Fisher
		Iris dataset. Left: Solution of \eqref{eq-adv1_cvc} with budget $|\varepsilon|\le 7$.
		Right: Solution of \eqref{eq-adv2_cvc} for a given deviation level $\delta_0=1200$.}
	\label{fig:iris-BL}
\end{figure}

	To analyze the bilevel models \eqref{eq-adv1_cvc} and \eqref{eq-adv2_cvc}, we fix a
perturbation budget $a=7$ and a target deviation level $\delta_0=1200$ in \eqref{eq-adv1_cvc} and \eqref{eq-adv2_cvc}, respectively. The results are demonstrated in  
Figure~\ref{fig:iris-BL}. 
On the left, the star indicates at
$\varepsilon=-6.45$, $\delta(\varepsilon)$ attains its maximum value of \eqref{eq-adv1_cvc}, which is 4974. 
To solve \eqref{eq-adv2_cvc}, we fix a threshold $\delta_0=1200$. 
The star highlights the optimal value of \eqref{eq-adv2_cvc} is $\varepsilon^*=-5.65$. 
In other words, the smallest perturbation to  achieve the
prescribed effect level (1200) is $\varepsilon=-5.65$. 

Next, we will check the efficiency of the proposed bilevel models. We focus on solving \eqref{eq-adv1_cvc}. We apply \texttt{fmincon} in MATLAB to solve \eqref{eq-adv1_cvc} on the different UCI datasets (Fisher Iris, Seeds\footnote{We select one feature (Feature 1, that is, $d=1$) and set $\gamma=10$.} and Wine\footnote{We select one feature (Feature 13, that is, $d=1$) and set $\gamma=5$.}). 
In our implementation, we use the MATLAB command \texttt{fmincon} to solve \eqref{eq-adv1_cvc}, where
the interior-point algorithm is used in \texttt{fmincon} and  the gradient is approximated by forward finite differences.
We also reported the optimal solution obtained by direct method , i.e. evaluate $U(\varepsilon)$ at $\varepsilon\in[-a: \mathrm{step}: a]$ and return the maximum value. For the Fisher Iris dataset, the related parameters are the same as in Section~\ref{subsec:delta-discussion}. For the Seeds dataset, we set $a=2$, $\mathrm{step}=0.01$, and for the Wine dataset, we set $a=0.2$, $\mathrm{step}=0.001$. In all cases, the resulting additive perturbations are
chosen so that they do not exceed roughly $30\%$ of the corresponding feature range. 
The results are summarized in Table~\ref{tab:uci-grid-vs-fmincon}. 
\begin{table}[H]
\centering
\caption{Comparison between direct method and \texttt{fmincon}
	on different UCI datasets.}
\label{tab:uci-grid-vs-fmincon}
\begin{tabular}{l l c c c c c c c c }
	\toprule
	Dataset & Method 
	& $n$ 
	& $K$
	& $\varepsilon$ 
	& $U(\varepsilon)$ 
	& $N_{\mathrm{chg}}$ 
	& time [s] \\
		\midrule
	\multirow{2}{*}{Fisher Iris} 
	& direct method  &100   & 2   &-6.45  &4974   &52 &12.90      \\
	& \texttt{fmincon} &100 & 2 &-7.9700  &4470  &40  &2.202     \\
	\midrule
	\multirow{2}{*}{Seeds} 
	& direct method  & 140  & 2 &0.95   &5750   &25   &16.49     \\
	& \texttt{fmincon} & 140  & 2 &1.9850   &4752   &75   &4.815     \\
	\midrule
	\multirow{2}{*}{Wine} 
	& direct method  &178   & 3   &-0.154  &7422 &115   &13.51      \\
	& \texttt{fmincon} &178   & 3  &0.0387   &7316   &116 &4.562    \\
	\bottomrule
\end{tabular}
\end{table}
Table~\ref{tab:uci-grid-vs-fmincon} compares the direct method and \texttt{fmincon} on the different UCI datasets. For all datasets, the direct method consistently finds perturbations with larger deviation values $U(\varepsilon)$, but at the price of higher CPU time. In comparison, the perturbation $\varepsilon$ returned by \texttt{fmincon} is different from that by the direct method. The reason is that \texttt{fmincon} returns a stationary point of the constrained optimization problem \eqref{eq-adv1_cvc}.

Below we analyze the computational complexity of the algorithm. 
	In our bilevel model \eqref{eq-adv1_cvc}, the main computational cost comes from the lower-level
	convex clustering problems solved by Ssnal, while the evaluation of the deviation
	functions $U(\varepsilon)$ only requires simple counting and basic calculation and is therefore negligible.

\subsection{Other Measurements for Deviation Functions}
\label{subsec:other-dev-measures}
Notice that in clustering methods, there are many measurements
 that can be used to evaluate the clustering results, such as Rand index (RI) and Normalized Mutual Information (NMI). In this part we analyse the behaviour of RI and NMI under perturbations.

 RI and NMI are computed in the following way \cite{wang2025euclidean}. Let 
 	$
 	c_i^0 := \mathrm{cluster\_id}_0(i)
 	\ \text{and}\ 
 	c_i^{\varepsilon} := \mathrm{cluster\_id}_\varepsilon(i),
 	\ i=1,\dots,n,
 	$
 	denote the cluster labels of point $i$ in the baseline and perturbed
 	clusterings, respectively.  We consider all pairs
 	$\{i,j\}$ with $1\le i<j\le n$ and count $	b_1 := 
 	\bigl|\{(i,j): c_i^0 = c_j^0,\; c_i^{\varepsilon} = c_j^{\varepsilon}\}\bigr|$, $b_2 := 
 	\bigl|\{(i,j): c_i^0 \neq c_j^0,\; c_i^{\varepsilon} \neq c_j^{\varepsilon}\}\bigr|$. 
 	Here $b_1$ and $b_2$ are the numbers of pairs on which the two
 	clusterings both in the same cluster or both in different
 	clusters.  The
 	RI at $\varepsilon$ is defined as
 	\begin{equation*}
 		\mathrm{RI}
 		:= \frac{b_1 + b_2}{\frac{n(n-1)}{2}} \in [0,1].
 	\end{equation*}
 	Thus $\mathrm{RI}=1$ if and only if
 	$c_i^{\varepsilon} = c_i^0$ for all $i$, i.e., the two clusterings coincide, and
 	the index decreases towards $0$ as more pairs of points are assigned
 	differently.

 	To define NMI, recall $D^\mathrm{CVC}_{Y^*}$ in \eqref{eq-dmap} and
 	let $D^\mathrm{CVC}_{Y^*(\varepsilon)}=\{ \widehat{V}_1, \cdots, \widehat{V}_{\hat{K}}\}$ as two partitions of the $n$ points.  NMI is calculated by
 	\[
 	\mathrm{NMI}
 	:=
 	\frac{
 		\displaystyle
 		\sum_{i=1}^{\hat K} \sum_{j=1}^{K}
 		\bigl|\hat V_i \cap V_j\bigr|\,
 		\log\!\left(
 		\frac{n \, \bigl|\hat V_i \cap V_j\bigr|}{|\hat V_i| |V_j|}
 		\right)
 	}{
 		\displaystyle
 		\sqrt{
 			\left(
 			\sum_{i=1}^{\hat K} |\hat V_i| \log\!\left(\frac{|\hat V_i|}{n}\right)
 			\right)
 			\left(
 			\sum_{j=1}^{K} |V_j| \log\!\left(\frac{|V_j|}{n}\right)
 			\right)
 		}
 	}\in [0,1].
 	\]
 	$\mathrm{NMI}=1$ if and only if
 	$D^\mathrm{CVC}_{Y^*}$ and $D^\mathrm{CVC}_{Y^*(\varepsilon)}$ induce
 	exactly the same partition, while smaller values indicate that the two clusterings share less information.

\begin{figure}[H]
	\centering
	\begin{minipage}{0.48\linewidth}
		\centering
		\includegraphics[width=\linewidth]{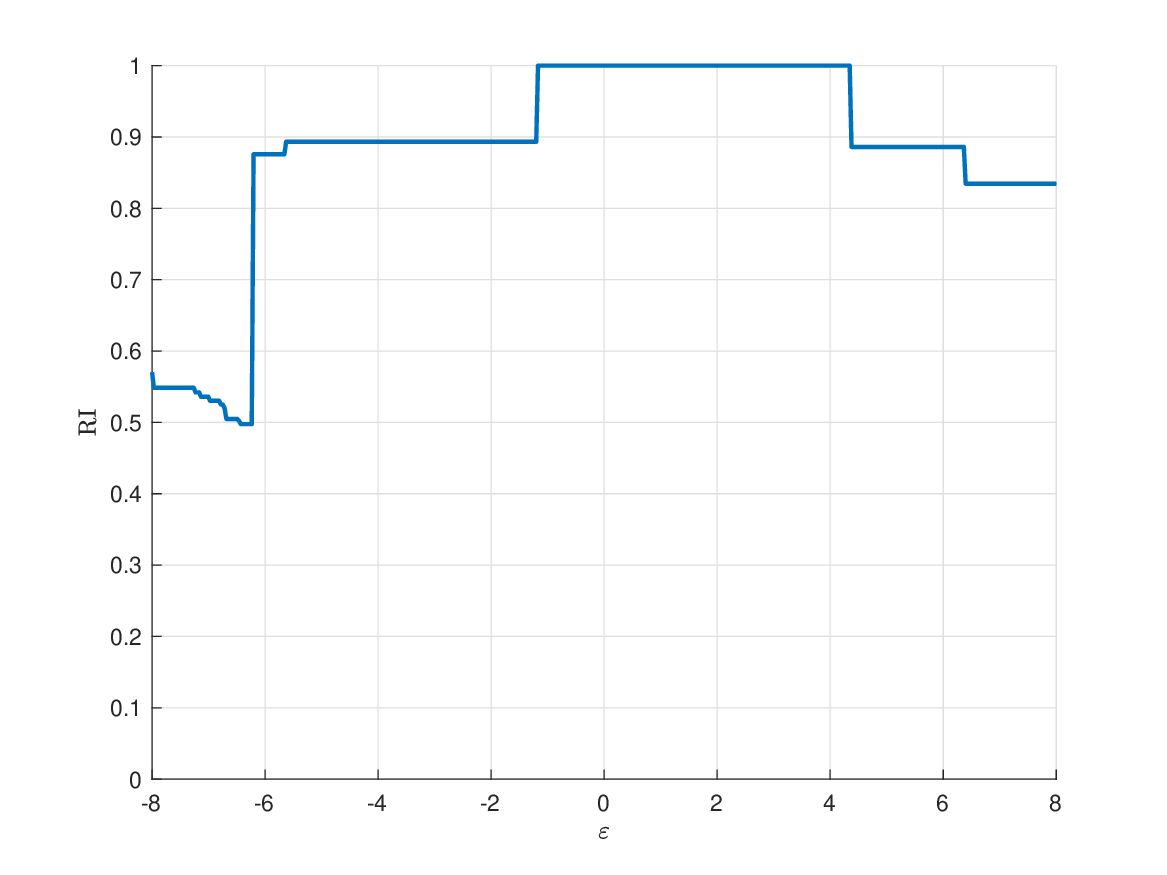}
	\end{minipage}\hfill
	\begin{minipage}{0.48\linewidth}
		\centering
		\includegraphics[width=\linewidth]{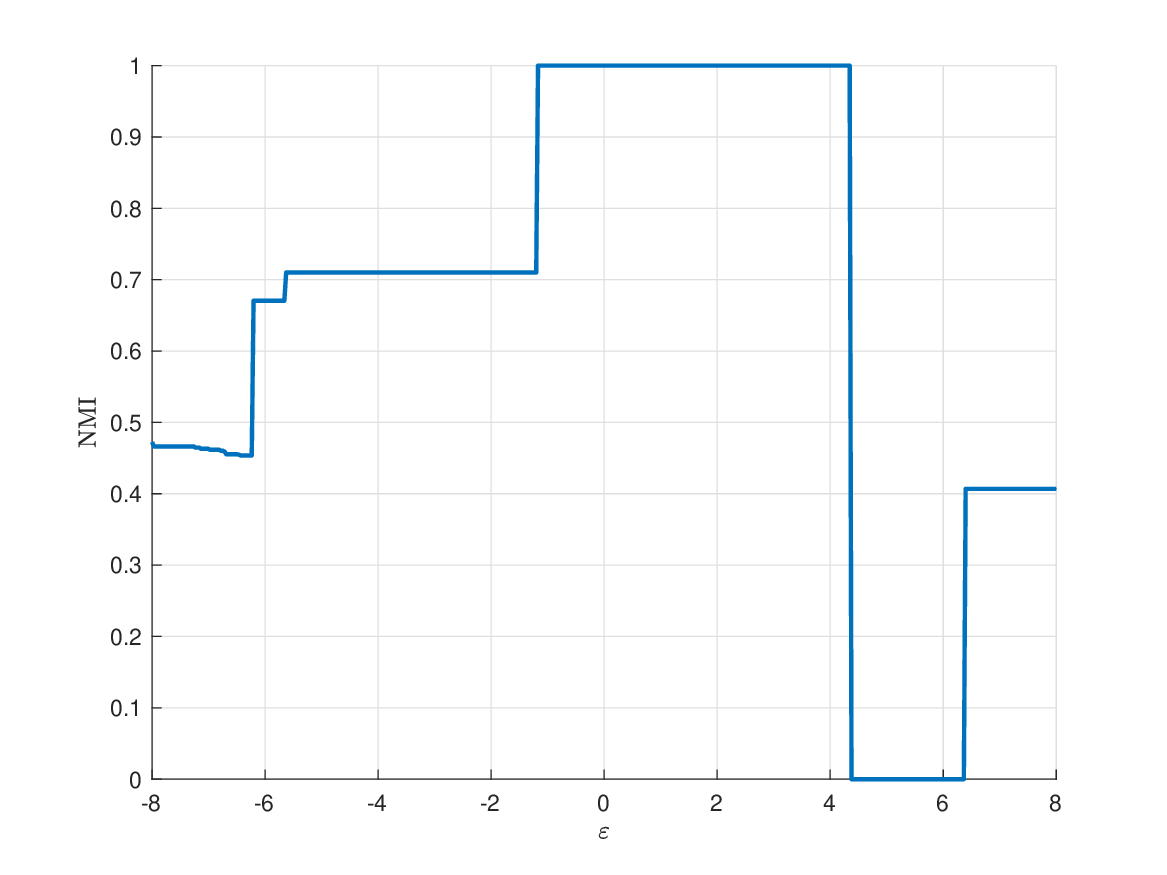}
	\end{minipage}
	\caption{
		RI and NMI on Fisher Iris dataset.
		Left: RI.
		Right: NMI.
	}
	\label{fig:iris-e-rinmi}
\end{figure}

In our test, the dataset and related parameters are the same as Section~\ref{subsec:robustness}. The results are demonstrated in Figure~\ref{fig:iris-e-rinmi}.
As we can see, as $|\varepsilon|$ increases, RI decreases, meaning that the clustering result becomes worse as the scale of noise grows.
Therefore, based on our deviation function, one can choose the  deviation function as
\begin{equation}\label{eq-ri}
	U^{\mathrm{RI}}(\varepsilon) := 1 - \mathrm{RI}(\varepsilon),
\end{equation}
which then satisfies the properties of a deviation function.

However, for NMI, it first decreases as $|\varepsilon|$ increases, but when $\varepsilon$ exceeds a certain threshold it drops to 0 and then jumps to a larger value. 
Due to this jump, NMI does not preserve monotonicity in $|\varepsilon|$.
Therefore, defining a reasonable deviation function
based on NMI is not trivial.

\begin{figure}[H]
		\centering
		\includegraphics[width=0.7\textwidth]{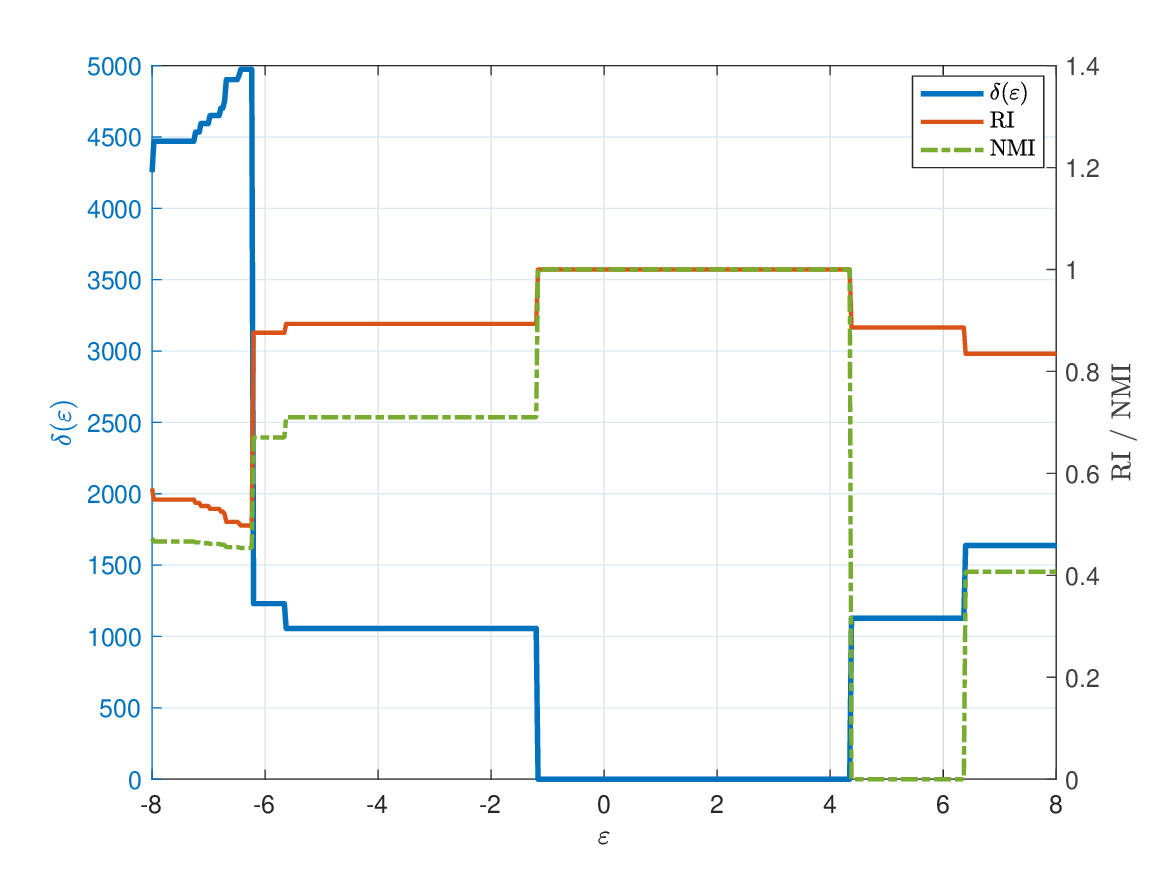}
		\caption{Deviation $\delta(\varepsilon)$, RI
			and NMI under the
			perturbation  $\varepsilon$ on Fisher Iris dataset.}
		\label{fig:iris-delta-ri-nmi}
	\end{figure}

Figure~\ref{fig:iris-delta-ri-nmi} overlays
	$\delta(\varepsilon)$, RI and NMI on the same plot.  This combined
	visualization clearly shows that intervals with  RI and NMI equal to 1 correspond to
	$\delta(\varepsilon)=0$, while large values of $\delta(\varepsilon)$ are
	accompanied by significantly reduced RI and NMI.  The three curves
	together therefore provide a consistent picture of clustering stability:
	the convex clustering model is robust with respect to moderate perturbations, but beyond certain thresholds the induced partition changes abruptly and the similarity indices drop accordingly. 

\subsection{Discussion and limitations}
The above experiments verify that the proposed bilevel models are numerically meaningful. 
Here we would like to highlight the following two important issues.
\begin{itemize}
	\item [(i)] The numerical experiments in Section~\ref{sec:numerics-bilevel} are carried out under an additive perturbation. 
	Nevertheless, one can also consider other types of perturbation (noises),  including multiplicative perturbations, feature-level perturbations, and graph-structured perturbations. 
	\item[(ii)] 
    The adversarial loss functions are very common  used in bilevel optimization. However, due to the limitation of space, we are not able to conduct detailed study on the property of adversarial loss functions. We will study and compare it with $\delta$-measure in our future research work. 
\end{itemize}

	\section{Conclusions}\label{sec:conclusions}
	In this paper, we proposed bilevel models for
	adversarial learning and instantiated it with the convex clustering model.
	Viewing attacks as data perturbations, we derived calmness-type results that connect the robustness of the learning model to properties of the solution mapping. 
	In particular, we identify the conditions for the robustness of convex clustering model. 
	We formulated two complementary bilevel models: model \eqref{eq-adv1}, which
	searches for worst-case perturbations under a norm budget, and model \eqref{eq-adv2}, which
	computes the smallest perturbation achieving a prescribed effect level.

	As a concrete deviation measure, we analyzed the $\delta$-measure, obtained explicit formulas for 2-way, 3-way and general $K$-way clustering, and identified cases in which $\delta$ behaves as a  deviation function. 
   The numerical
   results confirm that the proposed bilevel models are computationally viable
   and that the deviation measures, in particular the $\delta$-measure and the RI-based function as defined in \eqref{eq-ri}, provide a
   meaningful way to capture and quantify adversarial effects in
   convex clustering.

		However, several important questions remain open and deserve further investigation.
		On the one hand, it is still unclear how to design efficient algorithms for solving the two bilevel models for adversarial learning in more general settings.
		On the other hand,  all our experiments are conducted in a white-box setting.  In practice, the black-box clustering attacks are more prevalent. Therefore, how to design specified and efficient algorithms to solve the bilevel model is an interesting topic, which is worth further investigation.

\section*{Conflict of interest}
The authors declare that they have no conflict of interest.

\section*{Acknowledgments}
We would like to thank the three anonymous reviewers for their great comments based on which the paper has been significantly improved. We would also like to thank Jiani Li for her effort in identifying the definition of calmness.

\bibliographystyle{unsrt}
\bibliography{refs}
\end{document}